\documentclass[journal]{IEEEtran}
\usepackage{amsmath,amsfonts}
\usepackage{algorithmic}
\usepackage{algorithm}
\usepackage{array}
\usepackage{microtype}
\usepackage{textcomp}
\usepackage{stfloats}
\usepackage{url}
\usepackage{verbatim}
\usepackage{graphicx}
\usepackage{cite}
\usepackage{orcidlink}
\usepackage{longtable}
\usepackage{threeparttable}
\usepackage{multicol}
\usepackage{multirow}
\usepackage{booktabs}
\usepackage{siunitx}
\usepackage{enumitem}
\usepackage[hang,flushmargin,symbol]{footmisc}
\usepackage{hyperref}
\hypersetup{
  colorlinks,%
  citecolor=black,%
  filecolor=black,%
  linkcolor=black,%
  urlcolor=black
}
\usepackage{todonotes}
\usepackage{dirtree}

\newcolumntype{D}{p{2.8cm}}
\newcolumntype{T}{>{\centering\arraybackslash}m{1.2cm}}
\newcolumntype{C}{>{\centering\arraybackslash}m{1.9cm}}
\newcolumntype{L}{>{\centering\arraybackslash}m{2.75cm}}
\newcolumntype{Q}{>{\centering\arraybackslash}m{0.7cm}}
\newcolumntype{Z}{>{\centering\arraybackslash}m{0.5cm}}
\newcolumntype{P}{>{\centering\arraybackslash}m{1.7cm}}

\makeatletter
\if@twocolumn
  \renewcommand\normalsize{%
   \@setfontsize\normalsize\@xpt{12.5pt}%
   \abovedisplayskip=3 mm plus6pt minus 4pt
   \belowdisplayskip=3 mm plus6pt minus 4pt
   \abovedisplayshortskip=0.0 mm plus6pt
   \belowdisplayshortskip=2 mm plus4pt minus 4pt
   \let\@listi\@listI}%

  \renewcommand\small{%
   \@setfontsize\small{8.5pt}\@xpt
   \abovedisplayskip 8.5\p@ \@plus3\p@ \@minus4\p@
   \abovedisplayshortskip \z@ \@plus2\p@
   \belowdisplayshortskip 4\p@ \@plus2\p@ \@minus2\p@
   \def\@listi{\leftmargin\leftmargini
               \parsep 0\p@ \@plus1\p@ \@minus\p@
               \topsep 4\p@ \@plus2\p@ \@minus4\p@
               \itemsep0\p@}%
   \belowdisplayskip \abovedisplayskip}

\else
  \if@smallext
   \renewcommand\normalsize{%
   \@setfontsize\normalsize\@xpt\@xiipt
   \abovedisplayskip=3 mm plus6pt minus 4pt
   \belowdisplayskip=3 mm plus6pt minus 4pt
   \abovedisplayshortskip=0.0 mm plus6pt
   \belowdisplayshortskip=2 mm plus4pt minus 4pt
   \let\@listi\@listI}%

  \renewcommand\small{%
   \@setfontsize\small\@viiipt{9.5pt}%
   \abovedisplayskip 8.5\p@ \@plus3\p@ \@minus4\p@
   \abovedisplayshortskip \z@ \@plus2\p@
   \belowdisplayshortskip 4\p@ \@plus2\p@ \@minus2\p@
   \def\@listi{\leftmargin\leftmargini
               \parsep 0\p@ \@plus1\p@ \@minus\p@
               \topsep 4\p@ \@plus2\p@ \@minus4\p@
               \itemsep0\p@}%
   \belowdisplayskip \abovedisplayskip}
 \else
  \renewcommand\normalsize{%
   \@setfontsize\normalsize{9.5pt}{11.5pt}%
   \abovedisplayskip=3 mm plus6pt minus 4pt
   \belowdisplayskip=3 mm plus6pt minus 4pt
   \abovedisplayshortskip=0.0 mm plus6pt
   \belowdisplayshortskip=2 mm plus4pt minus 4pt
   \let\@listi\@listI}%

  \renewcommand\small{%
   \@setfontsize\small\@viiipt{9.25pt}%
   \abovedisplayskip 8.5\p@ \@plus3\p@ \@minus4\p@
   \abovedisplayshortskip \z@ \@plus2\p@
   \belowdisplayshortskip 4\p@ \@plus2\p@ \@minus2\p@
   \def\@listi{\leftmargin\leftmargini
               \parsep 0\p@ \@plus1\p@ \@minus\p@
               \topsep 4\p@ \@plus2\p@ \@minus4\p@
               \itemsep0\p@}%
   \belowdisplayskip \abovedisplayskip}
  \fi
\let\footnotesize\small
\makeatother

\usepackage{soul}
\soulregister{\cite}{7}
\soulregister{\ref}{7}
\soulregister{\SI}{7}
\newcommand{\highlight}[1]{\hl{#1}}
\renewcommand{\highlight}[1]{#1}

\begin{document}

\title{ROVER: A Multi-Season Dataset for Visual SLAM}

\author{Fabian Schmidt \orcidlink{0000-0003-3958-8932},
\highlight{Julian Daubermann, 
Marcel Mitschke,}
Constantin Blessing \orcidlink{0000-0005-8516-0269}, 
\highlight{Stephan Meyer,}\\
Markus Enzweiler \orcidlink{0000-0001-9211-9882}, 
and Abhinav Valada \orcidlink{0000-0003-4710-3114}
        
\thanks{Fabian Schmidt is with the Institute for Intelligent Systems, Esslingen University of Applied Sciences, Esslingen, Germany, and with the Department of Computer Science, University of Freiburg, Freiburg, Germany (E-mail: fabian.schmidt@hs-esslingen.de)}
\thanks{\highlight{Julian Daubermann, Marcel Mitschke, and Stephan Meyer are with ANDREAS STIHL AG \& Co. KG, Waiblingen, Germany (e-mail: julian.daubermann@stihl.de, marcel.mitschke@stihl.de, stephan.meyer@stihl.de)}}
\thanks{Constantin Blessing and Markus Enzweiler are with the Institute for Intelligent Systems, Esslingen University of Applied Sciences, Esslingen, Germany (E-mail: \{constantin.blessing, markus.enzweiler\}@hs-esslingen.de)}
\thanks{Abhinav Valada is with the Department of Computer Science, University of Freiburg, Freiburg, Germany (E-mail: valada@cs.uni-freiburg.de)}
}

\markboth{IEEE TRANSACTIONS ON ROBOTICS - Special COLLECTION: Visual SLAM. Preprint Version.}%
{Schmidt \MakeLowercase{\textit{et al.}}: ROVER - A Multi-Season Dataset for Visual SLAM}


\maketitle

\begin{abstract}

Robust Simultaneous Localization and Mapping (SLAM) is a crucial enabler for autonomous navigation in natural, \highlight{semi-structured} environments such as parks and gardens. However, these environments present unique challenges for SLAM due to frequent seasonal changes, varying light conditions, and dense vegetation. These factors often degrade the performance of visual SLAM algorithms originally developed for structured urban environments. To address this gap, we present ROVER, a comprehensive benchmark dataset tailored for evaluating visual SLAM algorithms under diverse environmental conditions and spatial configurations. We captured the dataset with a robotic platform equipped with monocular, stereo, and RGBD cameras, as well as inertial sensors. It covers 39 recordings across five outdoor locations, collected through all seasons and various lighting scenarios, i.e., day, dusk, and night with and without external lighting. With this novel dataset, we evaluate several traditional and deep learning-based SLAM methods and study their performance in diverse challenging conditions. The results demonstrate that while stereo-inertial and RGBD configurations generally perform better under favorable lighting and moderate vegetation, most SLAM systems perform poorly in low-light and high-vegetation scenarios, particularly during summer and autumn. Our analysis highlights the need for improved adaptability in visual SLAM algorithms for outdoor applications, as current systems struggle with dynamic environmental factors affecting scale, feature extraction, and trajectory consistency. This dataset provides a solid foundation for advancing visual SLAM research in real-world, \highlight{semi-structured} environments, fostering the development of more resilient SLAM systems for long-term outdoor localization and mapping. 
The dataset and the code of the benchmark are available under \url{https://iis-esslingen.github.io/rover}.
\end{abstract}


\section{Introduction}

\IEEEPARstart{S}{imultaneous} Localization and Mapping (SLAM) is a foundational technology in mobile robotics, enabling autonomous systems to construct a map of an unknown environment while simultaneously determining their location within it. This capability is crucial for robots to navigate effectively in dynamic, unstructured environments where external positioning systems, such as Global Navigation Satellite Systems (GNSS), may be unreliable or unavailable. In scenarios such as precision agriculture, forestry, and rugged outdoor environments, high-precision localization is essential for automation, particularly under challenging and constantly changing environmental conditions~\cite{ding2022agri_survey}.

\begin{figure}[t!]
\centering
\includegraphics[width=\linewidth]{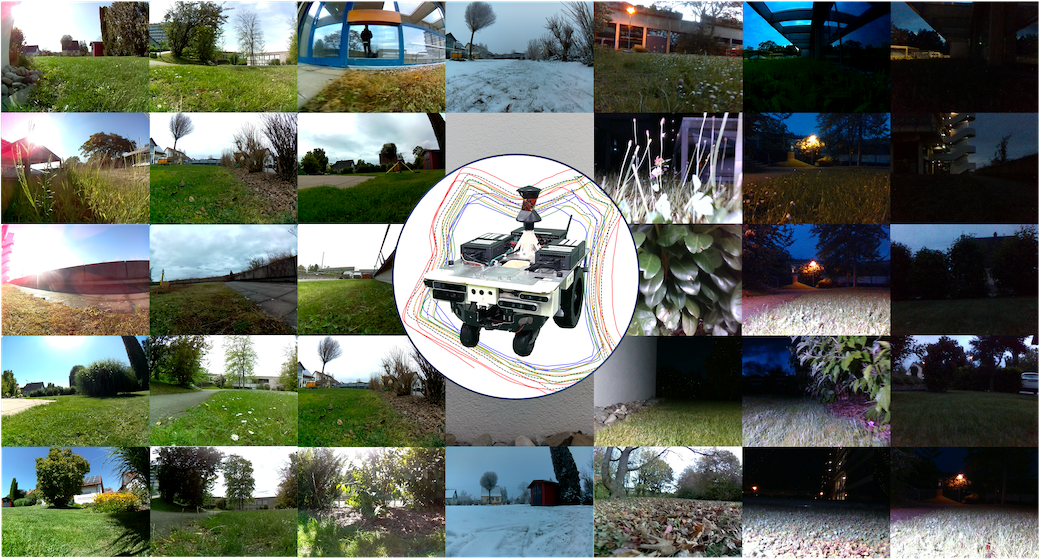}
\caption{Illustration of diverse outdoor environments captured across various seasons and lighting conditions. The scenes present natural challenges such as direct sunlight, low-texture areas, and reflective surfaces.}
\label{fig:cover_fig}
\end{figure}

Visual SLAM, in particular, has emerged as a key approach due to its ability to leverage visual data from cameras, making it a cost-effective and versatile solution~\cite{macario2022comprehensive}. However, most existing visual SLAM datasets are centered around structured environments such as indoor spaces~\cite{burri2016euroc} or urban areas~\cite{geiger2012kitti}, which lack the unique challenges presented by natural outdoor environments. Natural settings introduce a wide range of unpredictable factors, seasonal variations, fluctuating daylight, dynamic weather, and uneven terrain that affect long-term, robust localization and mapping. Visual SLAM systems must be resilient to both static and dynamic changes in these environments to achieve reliable performance over time~\cite{bevsic2022dynamic}.
\highlight{While certain existing datasets capture multi-season data in agricultural contexts (e.g., \cite{cuaran2023terrasentia, zujevs2021agriebv, crocetti2023ard-vo, polvara2022agri-robotic, marzoa2024magro, polvara2023blt}) or forest and off-road environments (e.g., \cite{boxan2024fomo, mactavish2018utias}), few focus on semi-structured, garden-like outdoor settings. These environments, such as parks or suburban lawns, introduce a distinct mix of man-made structures and natural features, along with moderate terrain variations, dense vegetation, and varying lighting conditions.} Effective assessment of environmental impacts on localization and mapping, especially for multi-session mapping across varying conditions, remains underexplored and is crucial for advancing visual SLAM systems~\cite{tourani2022visual}. \highlight{Consequently, there remains a notable gap in visual SLAM research for diverse camera setups (monocular, stereo, RGBD) under long-term seasonal and lighting changes in such semi-structured scenarios.} 

\highlight{To address this gap,} we introduce \textbf{ROVER} (\textbf{R}obot \textbf{O}utdoor \textbf{V}isual SLAM Dataset for \textbf{E}nvironmental \textbf{R}obustness), a millimeter-accurate visual SLAM benchmark dataset specifically \highlight{tailored for these types of park- and garden-like} outdoor environments. The dataset consists of over 39 recordings, covering \SI{7.2}{\kilo\meter} of trajectories in \SI{311}{\minute} and contains approximately 1~TB of raw data. It spans diverse semi-structured outdoor settings, including gardens, campus areas, and park-like scenarios, each with unique characteristics such as vegetation density, terrain slopes, and textureless or reflective surfaces. The data encompasses all four seasons and various lighting conditions, from daylight to night with external lighting, allowing a comprehensive evaluation of visual SLAM systems under dynamic conditions.
The dataset was collected using a \highlight{small-scaled UGV} equipped with a multi-sensor setup, incorporating monocular, stereo, and RGBD cameras as well as \highlight{camera-}internal and external inertial measurement units (IMU). A millimeter-accurate total station was used to capture the ground truth position of the robot. Designed to support both visual and visual-inertial SLAM evaluations, the dataset emphasizes \highlight{semi-structured} environmental challenges and long-term mapping consistency. It provides benchmarks for multi-session mapping, loop closure, and handling variations in seasonality and lighting, making it a valuable resource for developing more robust and adaptable SLAM systems for natural outdoor environments.

In addition to the dataset, this work provides a benchmark of traditional and deep learning-based visual SLAM methods, evaluating their performance across these challenging outdoor environments. These methods are applied across various sensor modalities, utilizing monocular, stereo, and RGBD approaches in combination with inertial data, allowing for a comprehensive comparison between traditional visual SLAM algorithms and data-driven deep learning approaches. The goal is to provide a detailed evaluation of how well these methods perform in \highlight{semi-structured} outdoor environments that pose significant challenges for localization and mapping.

The key contributions of this work are as follows:
\begin{itemize}[topsep=0pt,noitemsep]
    \item A millimeter-accurate localization benchmark dataset capturing extensive multi-session trajectories across diverse outdoor environments, utilizing a multi-sensor platform with various camera types and IMUs for robust evaluation of different SLAM configurations.
    \item Comprehensive coverage of multi-season, multi-weather, and multi-light conditions, addressing the challenges of long-term localization in dynamic environments.
    \item A benchmark comparing traditional and deep learning-based SLAM methods, providing insights into their performance under real-world outdoor conditions.
\end{itemize}

The remainder of this paper is structured as follows: Section~\ref{sec:relatedwork} reviews outdoor datasets relevant to visual and visual-inertial SLAM, focusing on natural and semi-structured environments. Section~\ref{datarecording} details the recording system, including the robotic platform, sensors, and data synchronization. Section~\ref{sec:rover} provides an overview of the dataset, describing the recording locations and detailing the characteristics of the recordings. Section~\ref{sec:benchmark} introduces the Visual SLAM benchmark, presenting experiments and results. Finally, Section~\ref{sec:conclusions} concludes with key findings and future research directions.

\section{Related Work}
\label{sec:relatedwork}

There are a wide variety of SLAM algorithms, each tailored to different applications, environments, and sensor setups. We do not provide a detailed discussion of individual algorithms but refer readers to comprehensive review papers that systematically cover the range of SLAM approaches~\cite{tourani2022visual, mokssit2023deep, tosi2024nerfs}. The specific methods used for the benchmark in this study are discussed in detail in Section~\ref{sec:slam_methods}.

To support the development of SLAM algorithms, diverse datasets are essential to capture the variety of conditions these systems must handle across different applications. SLAM methods often vary in their approach, depending on the target application, environment, and sensor setup, each requiring a unique range of data for thorough evaluation. In this study, we focus on datasets that capture natural settings using wheeled or legged robots, cars, or handheld devices, as these environments present unique challenges such as seasonal variation, dynamic weather, and complex lighting scenarios. \highlight{Note that numerous large-scale autonomous driving datasets also address seasonal changes in urban environments \cite{maddern20171oxford, larsson2019crossseason, yan2020eulongterm, wenzel20214seasons, burnett2023boreas}, but we do not discuss them in detail here as our emphasis is on unstructured and semi-structured, natural outdoor settings that are less represented in these urban-focused datasets.}
In the following sections, we discuss related outdoor datasets that provide diverse evaluation conditions for SLAM. These include datasets focused on both structured and unstructured settings such as urban and forested areas, as well as agricultural and garden environments. Table~\ref{table:related_datasets} provides an overview of these related datasets, detailing each dataset’s platform, environment type, diversity of scenarios, sensor configuration, and availability of ground truth data. This table also highlights how our dataset differs by focusing specifically on various natural outdoor settings that are underrepresented in existing datasets.

\begin{table*}
\centering
\caption{Comparison of related datasets: Platforms, environments, diversity, and sensor coverage.}
\fontsize{8pt}{8pt}\selectfont
\setlength\extrarowheight{1mm}
\setlength{\tabcolsep}{5.75pt}
\resizebox{\textwidth}{!}{%
\begin{threeparttable}
	\begin{tabular*}{1.1\linewidth}{D P L T Q Q Q Z Z Z Z L}
		\toprule
		  \multirow{ 2}{*}{\textbf{Dataset}} & \multirow{ 2}{*}{\textbf{Platform}} & \multirow{ 2}{*}{\textbf{Environment}} & \textbf{Consistent} & \multicolumn{3}{c}{\textbf{Diversity}} & \multicolumn{4}{c}{\textbf{Sensor}} & \multirow{ 2}{*}{\textbf{Ground Truth}} \\ 
		\cmidrule(lr){5-7}\cmidrule(lr){8-11} &  &  & \textbf{Trajectory} &                              
        \textbf{Weather} & \textbf{Lighting} & \textbf{Season} & \textbf{\highlight{Mono}} & \textbf{Stereo} & \textbf{RGBD} & \textbf{IMU} & \\ \toprule
        \multicolumn{12}{l}{\textbf{Heterogeneous Environments Datasets}} \\ \midrule
        SubT-MRS \cite{zhao2024subt} & ATV, Legged, UAV, UGV & Indoor, Campus, Off-Road, Underground &  & x & x &  & x &  &  & x & Laser Scanner + SLAM \\ 
        M3ED \cite{chaney2023m3ed} & Car, Legged, UAV & Forest, Urban &  &  &  &  & x &  &  & x & RTK-GNSS \\ 
        GEODE \cite{chen2024geode} & Boat, Hand, UGV & Off-Road, Underground, Urban, Water & & & & &  & x &  & x & RTK-GNSS/INS, Laser Tracker, Motion Capture \\ 
        FusionPortable \cite{jiao2022fusionportable} & Hand, Legged, UGV & Campus, Indoor, Urban & (x) &  & x &  &  & x &  & x & RTK-GNSS, Laser Tracker \& Scanner \\ 
        FusionPortableV2 \cite{wei2024fusionportablev2} & Car, Hand, Legged, UGV & Campus, Park, Indoor, Rural, Urban &  &  & x &  &  & x &  & x & RTK-GNSS, Laser Tracker \& Scanner \\ 
        EnvoDat \cite{nwankwo2024envodat} & Legged & Park, Indoor, Rural, Underground, Urban &  & x & x &  & x & x & x & x & SLAM \\ 
        PanoraMIS \cite{benseddik2020panoramis} & UAV, UGV & Indoor, Park &  &  &  &  & x &  &  & x & GNSS \\ 
        ShanghaiTech \cite{xu2024shanghaitech} & UGV & Indoor, Campus, Park &  &  &  &  & x &  & x & x & RTK-GNSS \\ 
        TartanAir \cite{wang2020tartanair} & UAV, UGV & Indoor, Campus, Forest, Off-Road, Park, Rural, Underground, Urban &  & x & x & x & x & x &  &  & Simulation \\ \midrule 

        \multicolumn{12}{l}{\textbf{Forest and Off-Road Environment Datasets}} \\ \midrule
        Wild-Places \cite{knights2023wildplaces} & Hand & Forest & (x) &  & x & x & x &  &  &  & SLAM \\ 
        WildScenes \cite{vidanapathirana2023wildscenes} & Hand & Forest &  &  & x & x & x &  &  & x & SLAM \\ 
        GOOSE \cite{mortimer2024goose} & Car & Forest, Off-Road &  & x &  &  & x &  &  & x & RTK-GNSS/INS \\ 
        FinnForest \cite{ali2020finnforest} & Car & Forest & (x) & x & x & x & x & x &  & x & GNSS \\ 
        FoMo \cite{boxan2024fomo} & UGV & Forest & x & x & x & x &  & x &  & x & \highlight{PPK}-GNSS\\ 
        SFU Mountain \cite{bruce2015sfu} & UGV & Forest & x & x & x &  & x & x &  & x & GNSS\\ 
        OORD \cite{gadd2024oord} & Car & Forest, Off-Road & x & x & x &  & x &  &  & x & GNSS/INS \\ 
        TartanDrive \cite{triest2022tartandrive} & ATV & Forest, Off-Road &  &  &  &  &  & x &  & x & GNSS \\ 
        TartanDrive 2.0 \cite{sivaprakasam2024tartandrive2} & ATV & Forest, Off-Road & (x) & x & x &  &  & x &  & x & GNSS \\ 
        Rally Estonia \cite{tampuu2023rally_estonia} & Car & Forest, Rural & (x) & x & x & x & x &  &  &  & GNSS \\ 
        GROUNDED \cite{ort2023grounded} & Car & Forest, Rural & x & x &  &  & x &  &  &  & RTK-GNSS/INS \\ 
        RELLIS-3D \cite{jiang2021rellis} & UGV & Off-Road &  &  &  &  & x &  &  & x & GNSS/INS\\ 
        UTIAS Long-Term \cite{mactavish2018utias} & UGV & \highlight{Campus,} Off-Road & x & x & x & x &  & x &  & x & VT\&R \\ \midrule 

        \multicolumn{12}{l}{\textbf{Agricultural Environment Datasets}} \\ \midrule
        Rosario \cite{pire2019rosario} & UGV & Farmland &  &  &  &  &  & x &  & x & RTK-GNSS \\ 
        Terrasentia \cite{cuaran2023terrasentia} & UGV & Farmland & x & x &  & x &  & x &  & x & GNSS \\ 
        FieldSAFE \cite{kragh2017fieldsafe} & Tractor & Farmland &  &  &  &  & x & x &  & x & GNSS \\ 
        AgriEBV \cite{zujevs2021agriebv} & UGV & Farmland, Forest, Off-Road &  & x &  & x &  & x & x & x & SLAM \\ 
        ARD-VO \cite{crocetti2023ard-vo} & UGV & Plantation & x &  &  & x & x & x &  & x & RTK-GNSS \\ 
        CitrusFarm \cite{teng2023citrus} & UGV & Plantation &  &  & x &  & x & x &  & x & RTK-GNSS \\ 
        Agri-Robotic \cite{polvara2022agri-robotic} & UGV & Plantation &  & x &  & x &  & x &  & x & RTK-GNSS \\ 
        Magro \cite{marzoa2024magro} & UGV & Plantation & x & x &  & x &  & x & x & x & RTK-GNSS \\ 
        BLT \cite{polvara2023blt} & UGV & Plantation & x & x &  & x & x & x & x & x & RTK-GNSS \\ \midrule 

        \multicolumn{12}{l}{\textbf{Garden and Park Environment Datasets}} \\ \midrule
        DiTer \cite{jeong2024diter} & Legged & Forest, Park &  &  &  &  & x &  & x & x & GNSS \\ 
        DiTer++ \cite{kim2024diter++} & Legged & Forest, Park & x &  & x &  & x &  & x & x & RTK-GNSS \\ 
        Newer College \cite{ramezani2020newer_college} & Hand & Campus, Park &  &  &  &  &  &  & x & x & Laser Scanner + ICP \\ 
        Newer College MC \cite{zhang2021multicamera_newer_college} & Hand & Campus, Indoor, Park &  &  &  &  & x & x &  & x & Laser Scanner + ICP \\ 
        BotanicGarden \cite{liu2024botanicgarden} & UGV & Garden, Park & &  &  &  & x & x &  & x & Laser Scanner + ICP \\ 
        EDEN \cite{le2021eden} & UGV & Garden &  & x & x &  & x & x &  &  & Simulation\\ 
        TB-Places \cite{leyva2019tb-places} & UGV & Garden &  &  & x &  & x & x &  & x & Laser Tracker \\ \midrule
        
        \textbf{ROVER} (Ours) & \textbf{UGV} & \textbf{Campus, Garden, Park} & \textbf{x} & \textbf{x} & \textbf{x} & \textbf{x} & \textbf{x} & \textbf{x} & \textbf{x} & \textbf{x} & Laser Tracker\\
		\bottomrule
	\end{tabular*}%
    \begin{tablenotes}
    \vspace{0.5em}
    \fontsize{8pt}{8pt}\selectfont
        \item[] Abbreviations: RTK - Real-Time Kinematic, \highlight{PPK - Post-Processing Kinematic,} INS - Inertial Navigation System, VT\&R - Visual Teach and Repeat \cite{furgale2010vtr}, ICP - Iterative Closest Point \cite{besl1992icp}
    \end{tablenotes}
\end{threeparttable}
} 
\label{table:related_datasets}
\end{table*}

\subsection*{Heterogeneous Environments Datasets}
Heterogeneous environment datasets are crucial for evaluating SLAM algorithms across diverse scenarios, focusing on structured and unstructured terrains rather than different natural settings. These datasets typically support multiple platforms, such as ground vehicles, unmanned aerial vehicles (UAV), and handheld devices, facilitating a broader assessment of SLAM performance under various conditions and sensor configurations.

The SubT-MRS~\cite{zhao2024subt}, M3ED~\cite{chaney2023m3ed}, and GEODE~\cite{chen2024geode} datasets each address a variety of challenging environments using a range of robotic platforms, including aerial, wheeled, and legged robots, and GEODE further expands the scope by including off-road and water settings. EnvoDat~\cite{nwankwo2024envodat} employs a legged robot across park-like, indoor, underground, and urban scenarios, focusing on conditions like dynamic entities, varied illumination, fog, rain, and low visibility. FusionPortable~\cite{jiao2022fusionportable} and FusionPortableV2~\cite{wei2024fusionportablev2} extend each other’s scope across indoor, campus, and urban settings, supporting handheld devices, UGVs, and cars, with FusionPortableV2 adding larger-scale coverage. PanoraMIS~\cite{benseddik2020panoramis} uniquely utilizes panoramic cameras across various environments collected via wheeled, aerial, and industrial robots. The ShanghaiTech~\cite{xu2024shanghaitech} dataset spans both indoor and outdoor settings with a UGV, while TartanAir~\cite{wang2020tartanair} provides simulated large-scale environments with urban, rural, and natural scenes, introducing diverse challenges with its dynamic elements.

\subsection*{Forest and Off-Road Environment Datasets}
Datasets with a focus on natural, unstructured environments, such as forests and off-road terrain, are essential for the further development of SLAM algorithms in complex real-world environments, where challenges such as uneven terrain, dense vegetation, and dynamic lighting are prevalent.

The Wild-Places~\cite{knights2023wildplaces} and WildScenes~\cite{vidanapathirana2023wildscenes} datasets are both centered on forest environments, utilizing a handheld system with LiDAR, cameras, and IMU. WildScenes extends Wild-Places by adding high-resolution semantic annotations in 2D and dense 3D LiDAR point clouds, aiding in re-localization and loop closure detection across different seasons.
The GOOSE~\cite{mortimer2024goose} and FinnForest~\cite{ali2020finnforest} datasets both use a car platform to capture data in forest and off-road terrains. GOOSE focuses on perception under various weather conditions with semantic labels in 2D and 3D, while FinnForest collects data under different lighting and seasonal conditions, including day, night, summer, and winter, with partially repeated routes.
The FoMo dataset~\cite{boxan2024fomo} and SFU Mountain dataset~\cite{bruce2015sfu} both involve off-road environments, with FoMo using an ATV platform to capture seasonal changes in a forest and SFU Mountain using a UGV focusing on woodland trails under diverse lighting and weather conditions.
The OORD~\cite{gadd2024oord} and TartanDrive~\cite{triest2022tartandrive} datasets capture data in rugged off-road environments using different platforms: OORD emphasizes radar data for place recognition under adverse conditions in the Scottish Highlands, while TartanDrive and its follow-up, TartanDrive~2.0~\cite{sivaprakasam2024tartandrive2}, collect off-road driving data using an ATV with a variety of sensors, including LiDAR in the updated version.
The Rally Estonia~\cite{tampuu2023rally_estonia} and GROUNDED~\cite{ort2023grounded} datasets both focus on rural terrains, capturing data in forested areas using cars and emphasizing localization under varying weather conditions.
In contrast, RELLIS-3D~\cite{jiang2021rellis} provides detailed 2D and 3D semantic labels in off-road settings using a UGV, while UTIAS Long-Term~\cite{mactavish2018utias} focuses on autonomous, repeated route data in off-road environments, capturing full daily lighting cycles and seasonal changes.

\subsection*{Agricultural Environment Datasets}
Agricultural environment datasets are crucial for advancing SLAM algorithms in structured yet dynamic settings such as farmland and plantations, where terrain, crop types, and seasonal changes introduce unique challenges.

The FieldSAFE~\cite{kragh2017fieldsafe} and AgriEBV~\cite{zujevs2021agriebv} datasets capture data in farmland environments using UGV platforms with multiple sensors, including LiDAR, cameras, and IMUs, to support SLAM research under varied weather conditions and across different seasons. FieldSAFE is notable for incorporating dynamic obstacles, while AgriEBV extends to include off-road and forest areas as well.
The ARD-VO~\cite{crocetti2023ard-vo}, CitrusFarm~\cite{teng2023citrus}, and Agri-Robotic~\cite{polvara2022agri-robotic} datasets focus on plantation environments, with ARD-VO collecting data in olive groves and vineyards, CitrusFarm in citrus plantations using a wheeled robot with a variety of sensors, and Agri-Robotic documenting long-term changes in a vineyard to analyze how environmental variations impact mapping and localization tasks.
The Magro~\cite{marzoa2024magro} and BLT~\cite{polvara2023blt} datasets both capture data in orchards and vineyards over multiple sessions across different seasons. Magro focuses on an apple plantation with repeated trajectories under varying conditions, while BLT documents canopy growth in a vineyard over seven months, making them valuable for studying long-term mapping, localization, and loop closure in agricultural settings.

\subsection*{Garden and Park Environment Datasets}
Datasets focusing on garden and park environments are essential for evaluating SLAM algorithms in semi-structured outdoor settings. These environments typically feature diverse terrain types, vegetation, and man-made structures, offering a mix of natural and artificial elements that challenge visual and sensor-based navigation systems.

The DiTer~\cite{jeong2024diter} and DiTer++~\cite{kim2024diter++} datasets focus on park-like and forested areas, using legged robots to capture various terrains, including sandy roads and slopes. DiTer++ expands on the original dataset with multi-robot setups, heterogeneous LiDARs, and day-night sequences, making it more versatile for SLAM evaluation.
The Newer College~\cite{ramezani2020newer_college} and Newer College Multi-Camera~\cite{zhang2021multicamera_newer_college} datasets are collected on the campus of New College, Oxford. The original dataset, captured with a handheld system, includes structured buildings, vegetation, and open spaces, while the extended version introduces a multi-camera system with fisheye lenses, addressing challenges like abrupt lighting changes and textureless surfaces.
The BotanicGarden dataset~\cite{liu2024botanicgarden}, captured in a diverse botanical garden using a UGV platform, features thick woods, riversides, narrow trails, and open grasslands, providing varied scenarios for testing SLAM in natural and semi-structured environments.
The EDEN dataset~\cite{le2021eden} is a synthetic multimodal dataset with over 100 garden models, offering rich variations in landscapes and vegetation. It provides data under different weather and lighting conditions, inspired by the TB-Places dataset~\cite{leyva2019tb-places}, which is designed for visual place recognition in garden environments. TB-Places, captured with a lawn mower, includes diverse terrain types and obstacles, filling a critical gap in natural environment place recognition data.

\begin{figure*}[b!]
\centering
\includegraphics[width=\linewidth]{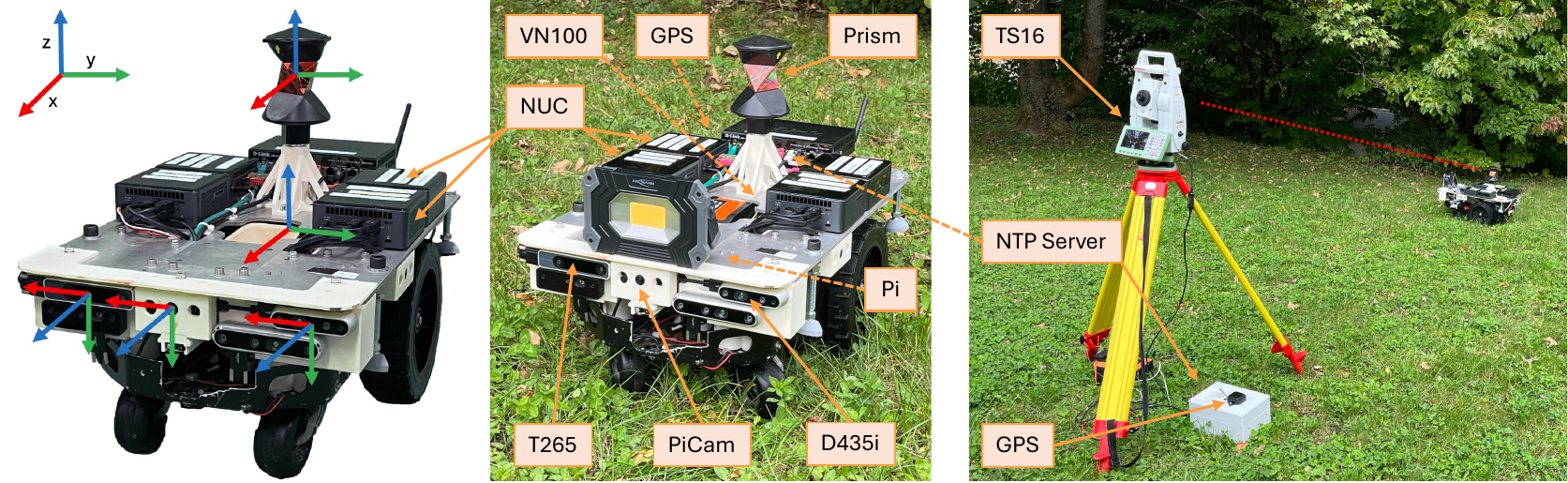}
\caption{Overview of the Robotic Platform. The left section shows the robotic platform, including specific sensor coordinate systems. The center illustrates the platform in an outdoor field environment, detailing the sensors and computing devices used to log and time-synchronize the sensor data. The right section presents the ground truth system in operation, with the robotic platform in the field. Components placed under the hood are indicated with a dashed line.}
\label{fig:robot_platform}
\end{figure*}
\begin{table*}[b!]
\centering
\caption{Overview of sensors and ground truth system with specifications.}
\setlength\extrarowheight{1mm}
\fontsize{8pt}{8pt}\selectfont
\resizebox{\textwidth}{!}{%
    \begin{tabular*}{1.1\textwidth}{@{\extracolsep{\fill}} llll} 
        \toprule
        \textbf{Sensor} & \textbf{Modality} & \textbf{Characteristics} & \textbf{Data Rate [Hz]} \\
        \toprule
        Joy IT Pi Camera & Camera & Resolution: 640 x 480 RGB, Rolling Shutter, FoV: \ang{160}/\ang{122}/\ang{89.5}, Wide Angle, Fixed Focus & 30 \\\midrule
        \multirow{4}{*}{Intel RealSense D435i} & \multirow{3}{*}{Camera} & Resolution: 640 x 480, Global Shutter & \multirow{3}{*}{30} \\ 
         & & RGB: FOV: \ang{77}/\ang{69}/\ang{42}, Rolling Shutter, Fixed Focus & \\
         & & Depth: Active IR Stereo, 0.2m - 10m, FoV: \ang{94}/\ang{86}/\ang{57} & \\ 
         & IMU & 6 DoF: Accelerometer, Gyrometer & 100, 200 \\ \midrule
        \multirow{2}{*}{Intel RealSense T265} & Camera & Resolution: 848 x 800 Monochrome, Global Shutter, FoV: \ang{160} Diagonal, Fisheye Lens, Fixed Focus & 30 \\ 
         & IMU & 6 DoF: Accelerometer, Gyrometer & 65, 200 \\ \midrule
        VectorNav VN100 & IMU & 9 DoF: Accelerometer, Gyrometer, Magnetometer & 65 \\ \midrule
        Leica TS16 Totalstation & Laser Tracker & 3 DoF Position, Accuracy: 1mm + 1,5 ppm & 2 - 5 \\
        \bottomrule
    \end{tabular*}%
} 
\label{table:sensors}
\end{table*}

\highlight{Although some recent multi-season datasets share similarities with our work, most notably FoMo and UTIAS Long-Term, they diverge in key aspects. FoMo captures stereo-only camera data in boreal forests using a large off-road UGV, including some nighttime sequences but lacking garden and park scenarios and RGBD sensing. UTIAS Long-Term also employs a large off-road capable UGV, gathering data that distinguishes lighting variations captured on campus from seasonal variations recorded in off-road environments from January to May, and it provides stereo keyframe images sampled at roughly every \SI{0.2}{\meter}. In contrast, our dataset covers both lighting and multi-season changes across five semi-structured suburban locations, spanning all four seasons, with monocular, stereo, and RGBD recordings from a small lawnmower-like UGV that is sensitive to uneven terrain. Existing works do not offer the same combination of multi-sensor modalities and year-round recordings in garden and park-like settings as our dataset. Many publicly available datasets still fall short in representing the full range of lighting, seasonal, and weather variations (see Table~\ref{table:related_datasets}).} Furthermore, there is a significant need for datasets that specifically assess the impact of these environmental factors on localization and mapping, particularly for visual SLAM systems using various camera types, such as monocular, stereo, and RGBD, with the potential for use in multi-session mapping studies. To address these gaps, our dataset offers a high-precision, long-term benchmark, capturing consistent routes across five distinct \highlight{semi-structured} outdoor environments under a wide range of conditions, including varied weather and lighting scenarios across all seasons. The dataset also addresses challenging real-world factors, such as dynamic illumination changes, reflective and textureless surfaces, motion blur, occlusions, and uneven terrain, providing a comprehensive resource for evaluating and advancing visual SLAM algorithms.

\section{Data Recording System} 
\label{datarecording}
The data recording system is designed to enable reliable, synchronized, and efficient data capture, which is crucial for evaluating SLAM algorithms across diverse environments. Built on a holonomic robotic platform, the system is equipped with a range of sensors, including cameras, IMUs, and a ground truth system. Each sensor is carefully selected and strategically positioned to ensure optimal data quality and comprehensive environmental coverage.
The following subsections detail the platform design, the sensor suite and its calibration, the dedicated computing architecture, and the time synchronization setup.

\subsection{Robotic Platform}
The robotic platform used for data collection is based on a prototype of a robotic lawn mower that has been modified to support a sensor suite and onboard computing resources mounted on top, as shown in Figure~\ref{fig:robot_platform}. The platform is driven by two large rear wheels, which provide the primary locomotion, while two smaller front wheels primarily serve to enhance stability. This drive configuration ensures efficient movement and control over the terrain. However, due to the design and weight distribution, the platform is highly sensitive to changes in surface conditions, which can result in vibrations and shaking when traversing uneven or rough terrain. This increased complexity highlights the challenges encountered in real-world robotic applications, where variable environments can affect both navigation and data quality.

The holonomic robotic platform is controlled using a game controller, allowing for manual navigation and precise handling of the robot's movement. While the platform is capable of reaching speeds up to \SI{1}{\meter\per\second}, it operates at a constant speed of \SI{0.5}{\meter\per\second} during all data acquisition sessions to maintain consistency and accuracy. This constant velocity minimizes the influence of dynamic movement on the sensors, ensuring uniformity in the recorded data across different runs and environmental conditions.
In addition, an external light source is mounted on the front of the sensor and computing platform to enable night-time data recording.

\subsection{Sensors}
The sensor suite on the robotic platform was carefully selected to enable robust data collection across multiple modalities, supporting the evaluation of SLAM systems in diverse environmental conditions. This suite includes cameras, IMUs, and a high-precision ground truth system, all of which are mounted strategically to maximize data accuracy and relevance. Table~\ref{table:sensors} provides an overview of the specific characteristics of each sensor, including their roles and specifications within the platform. The following subsections describe each sensor type in detail, covering the cameras, IMUs, ground truth system, and calibration process essential for accurate sensor fusion and reliable benchmarking.

\begin{figure*}[b!]
  \centering
  \footnotesize
  \setlength{\tabcolsep}{0.25cm}
      {\renewcommand{\arraystretch}{1}
        \begin{tabular}{p{5.9cm}p{5.9cm}p{5.9cm}}
            \includegraphics[height=4.7cm]{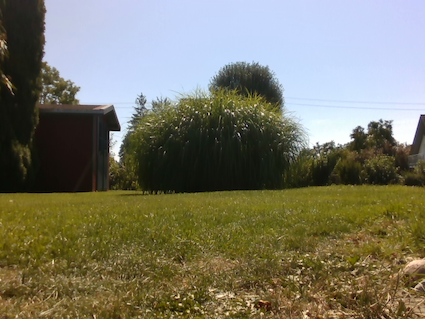} &
        \includegraphics[height=4.7cm]{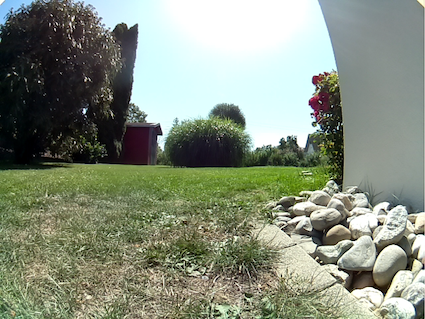} &
        \includegraphics[height=4.7cm]{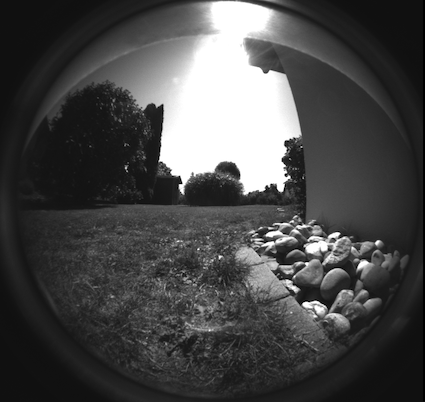} \\
        \multicolumn{1}{c}{(a) D435i} & \multicolumn{1}{c}{(b) PiCam} & \multicolumn{1}{c}{(c) T265} \\
    \end{tabular}}
    \caption{Camera Perspectives. Example images showcasing the perspectives from each camera on the robotic platform. (a) The D435i captures RGB and depth data with a standard wide-angle view. (b) The PiCam provides a broader, wide-angle RGB perspective. (c) The T265 offers a fisheye view from a monochrome stereo configuration, optimized for tracking.}
      \label{fig:camera_perspectives}
\end{figure*}

\subsubsection{Cameras}
Cameras are the central component of the dataset, providing diverse perspectives and depth of information. The cameras are mounted at the front of the sensor platform and include monocular, stereo, and~{RGBD} cameras, each selected to capture different environmental aspects, see Figure~\ref{fig:robot_platform}.

The Joy IT Pi Camera (PiCam) is a budget-friendly monocular camera that offers $640 \times 480$ RGB images with a wide \SI{160}{\degree} field of view. However, it uses a rolling shutter, which can result in motion artifacts.
The Intel RealSense D435i captures both RGB and depth data and includes an integrated IMU, enhancing its potential for motion tracking applications. Its depth sensor features a global shutter, reducing motion distortion and making it ideal for applications requiring accurate medium-distance depth measurements. The RGB sensor, however, uses a rolling shutter with a narrower field of view (FoV), balancing cost and functionality for basic visual data.
In contrast, the Intel RealSense T265 is designed specifically for visual-inertial tracking, offering a fisheye stereo configuration with a broad FoV and a global shutter that improves performance during fast motion. Combined with its built-in IMU, the T265 is highly effective for localization and motion tracking, though it only captures monochrome images, unlike the RGB data provided by the D435i and other cameras.

The choice of cameras reflects a balance between cost and functionality. Monocular cameras such as the PiCam provide standard RGB images, while stereo and RGBD cameras such as the D435i and T265 enhance depth perception and spatial awareness. These cameras also feature different fields of view, from wide-angle to fisheye lenses, to maximize environmental coverage. Figure~\ref{fig:camera_perspectives} highlights the diverse perspectives captured by each camera.


\subsubsection{IMUs}
The Intel Realsense D435i and T265 cameras feature internal IMUs that capture 6~DoF data using accelerometers and gyroscopes. Since these IMUs are integrated within the cameras, they are positioned at the front of the sensor platform, making them highly sensitive to uneven terrain and dynamic movements. In addition to the internal IMUs, the platform is equipped with a VectorNav VN100, a high-precision 9~DoF IMU mounted at the base of the robot. This IMU combines accelerometers, gyroscopes, and magnetometers to provide a complete understanding of the robot’s orientation and rotation along its axes. Its placement on the rotation axis of the body allows it to capture more accurate motion dynamics, particularly when the platform encounters uneven terrain.

\subsubsection{Ground Truth}
To ensure accurate 3~DoF position tracking, the Leica TS16 Totalstation is employed as the ground truth system. It captures the precise position of a prism mounted on the platform, providing millimeter-level precision. The system records data at a frequency of 2-\SI{5}{\hertz}, offering highly reliable reference data for validating sensor accuracy and positioning.

\begin{figure*}[b!]
\centering
\includegraphics[width=0.6\linewidth]{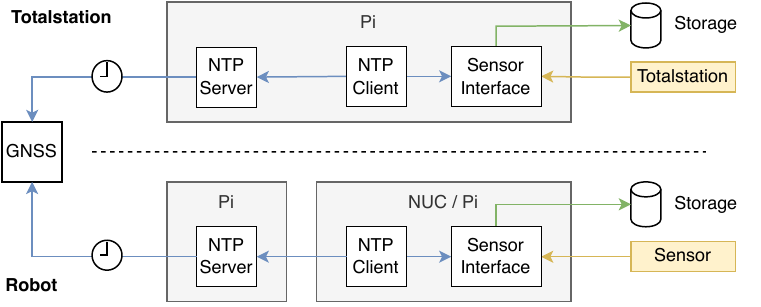}
\caption{The system architecture for data recording and synchronization. The diagram illustrates the setup for sensor data collection, processing, and time synchronization. The blue pathways represent time synchronization via GNSS and NTP protocols, ensuring that each device remains accurately aligned. The yellow pathways depict sensor data flow, which each computing device handles individually. Finally, the green pathways indicate the flow of time-synchronized data stored on external SSDs for the sensor platform and a dedicated storage device for ground truth data collection.}
\label{fig:recording_system}
\end{figure*}

\subsubsection{Calibration}
Accurate sensor calibration is crucial to ensure that the data collected from the robotic platform is reliable and can be properly fused across different modalities. The calibration process involved two primary methods: Kalibr toolbox~\cite{furgale2013kalibr} for camera and IMU calibration and a manual CAD-based approach for calibrating the cameras with respect to the prism used in the ground truth system.

For the calibration of the cameras and IMUs, we employed the Kalibr toolbox, a widely used tool for multi-sensor calibration in robotics. This process involved three key calibration steps. First, the intrinsic parameters of each camera, including focal length, optical center, and lens distortion coefficients, were calibrated using a checkerboard pattern. Second, the IMU biases and noise characteristics were calibrated, accounting for factors such as sensor drift over time. Finally, the extrinsic parameters between the cameras and IMUs were determined to synchronize the data for sensor fusion. This calibration step involved aligning the relative position and orientation of the IMU with respect to each camera, ensuring that the motion data from the IMU can be accurately fused with the visual data from the cameras.

To accurately integrate the sensor data with the ground truth system provided by the Totalstation, a CAD-based calibration method was employed to align the sensors with the prism mounted on the platform. Using the CAD model of the sensor platform, the relative position and orientation between the sensors and the prism were determined. This step is critical for ensuring that the ground truth measurements from the Totalstation align correctly with the sensor data, allowing for accurate validation of the platform's positional accuracy.

\subsection{System Overview}
The system architecture is designed to facilitate efficient data recording, storage, and time synchronization across all sensors and computing devices, enabling accurate and reliable SLAM evaluation. Figure~\ref{fig:recording_system} presents an overview of the platform's data recording and synchronization setup, which includes dedicated devices for sensor processing and an integrated time synchronization system to ensure temporal alignment across all recorded data. The following subsections detail the computing architecture and time synchronization methods employed to support seamless data capture and storage.

\subsubsection{Computing Architecture}
The computing architecture is designed to efficiently manage sensor data processing and storage, with each sensor assigned its own dedicated computing device. 
The system employs two types of computing devices: an Intel NUC10i5FNB (NUC) with 8GB of RAM and a Raspberry Pi 4 Model B Rev 1.4 (Pi) with 4GB of RAM. The NUCs handle data from the Intel RealSense D435i, Intel RealSense T265, and the VN100, while the Pi manages the PiCam. Each sensor is connected to its respective computing device via USB interfaces.
A sensor interface is responsible for collecting data from each sensor and storing it on an external SSD hard drive connected to the computing device. For recording the ground truth trajectory, a separate Pi is connected to the Totalstation via RS232, which manages the logging of ground truth data onto an external SSD.

\subsubsection{Time Synchronization}
The time synchronization across the entire system is crucial for ensuring the temporal alignment of data collected from the sensors and the ground truth system. This synchronization is achieved through a GNSS signal provided by a u-blox GPS receiver, which serves as the primary time reference for the system. Both Pi devices are connected to a u-blox C099-F9P Application Board, featuring a u-blox ZED-F9P-01B module that receives accurate GNSS signals.
On the sensor platform, one of the Pis serves as the NTP (Network Time Protocol) server and is connected via Ethernet to a network switch. This switch links the NTP server to the other computing devices, ensuring that all sensor computing units share a common time base. By leveraging the GNSS signal and NTP synchronization, the system achieves \highlight{millisecond}-level accuracy, allowing the NTP server to synchronize the NTP clients on these devices and provide precise timestamps for data collection.

\section{ROVER Dataset}
\label{sec:rover}

Designed to present a diverse set of environmental conditions, this dataset provides a comprehensive basis for evaluating the robustness of SLAM algorithms. With varied locations, seasonal shifts, and environmental factors such as weather and lighting, it enables a thorough assessment of SLAM adaptability in natural and semi-structured settings.
The following subsections detail the recording locations, their defining characteristics, and the methodology behind data collection, highlighting the conditions under which the data was acquired, and also discuss the organization of the dataset.

\subsection{Locations}

\begin{figure}[t!]
  \centering
  \footnotesize
  \setlength{\tabcolsep}{0.0cm}
  {\renewcommand{\arraystretch}{1}
    \begin{tabular}{p{\linewidth}}
        \includegraphics[width=\linewidth]{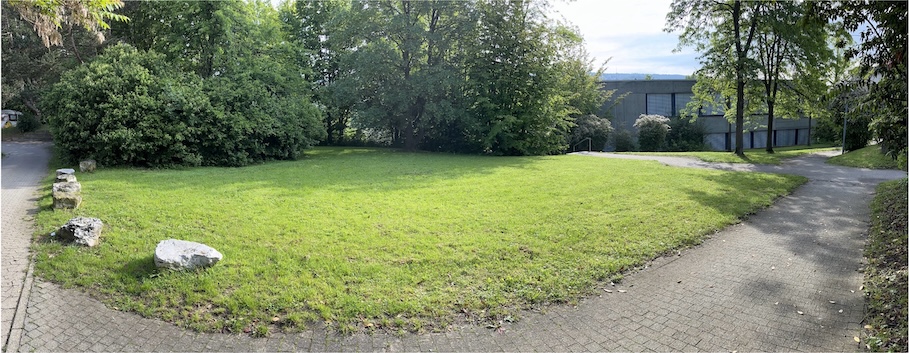} \\[0.5em]
        \multicolumn{1}{c}{(a) Park} \\[0.5em]
        
        \includegraphics[width=\linewidth]{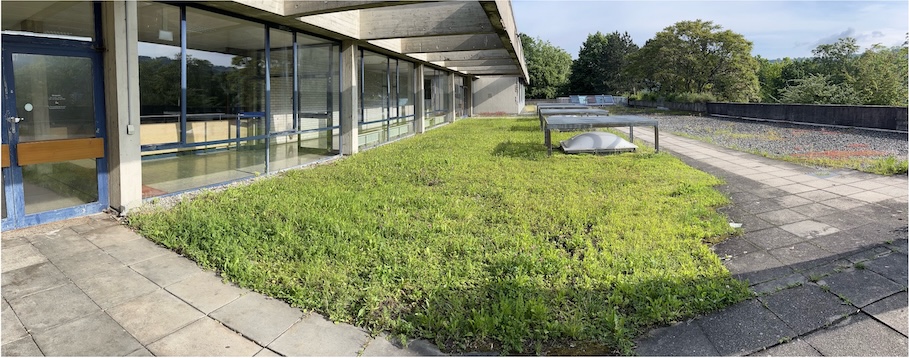} \\[0.5em]
        \multicolumn{1}{c}{(b) Campus Small} \\[0.5em]
        
        \includegraphics[width=\linewidth]{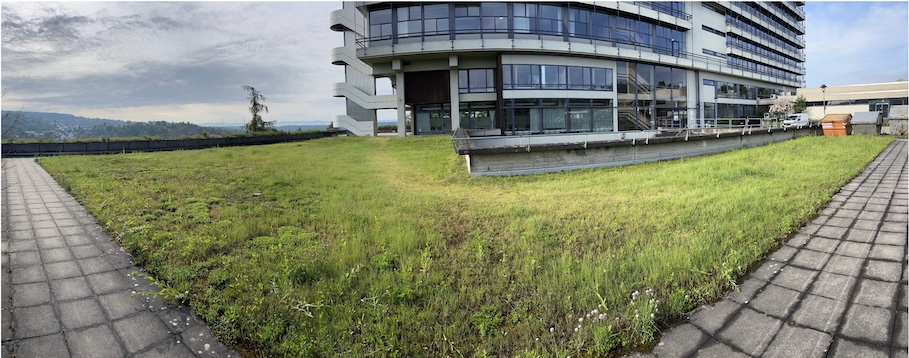} \\[0.5em]
        \multicolumn{1}{c}{(c) Campus Large} \\[0.5em]
        
        \includegraphics[width=\linewidth]{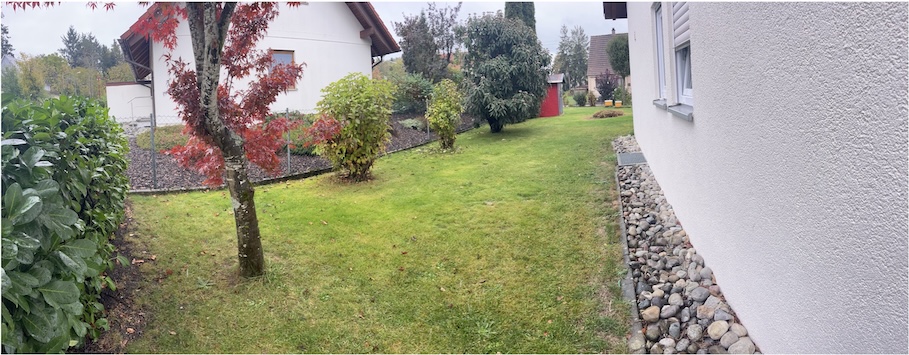} \\[0.5em]
        \multicolumn{1}{c}{(d) Garden Small} \\[0.5em]
        
        \includegraphics[width=\linewidth]{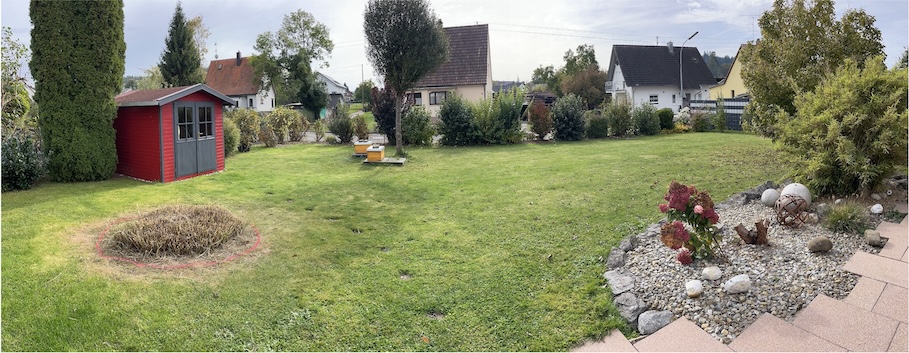} \\[0.5em]
        \multicolumn{1}{c}{(e) Garden Large} \\
    \end{tabular}}
    \caption{Overview of data recording locations. (a) Park: Natural area with moderate slopes and varied vegetation; (b) Campus Small: Compact lawn near buildings with low vegetation; (c) Campus Large: Rooftop with minimal vegetation and open views; (d) Garden Small: Shaded garden with moderate vegetation; (e) Garden Large: Traditional garden with diverse vegetation and open light.}
    \label{fig:environments}
\end{figure}

To capture a broad range of outdoor environments, a diverse set of locations was chosen for data recording. Each location presents unique characteristics and challenges, ensuring that the dataset reflects a variety of real-world conditions. 
The selection of locations was driven by several key criteria, which are critical for assessing the performance of the robotic platform in different contexts. These criteria include area size, shape complexity, slope inclination, long-term stability, vegetation, and brightness, as follows:
\begin{itemize}[noitemsep, topsep=0pt]
    \item \textit{Area Size:} The size of the lawn or area influences the complexity of navigation and path planning. Larger areas demand longer operational times and more sophisticated path strategies, while smaller areas focus on precision.
    
    \item \textit{Shape Complexity:} The shape of each location plays a significant role in the robot’s ability to navigate and localize itself. Simple geometric shapes, such as squares or rectangles, are easier to navigate, while more complex shapes introduce additional challenges due to frequent rotations and dynamic movements.

    \item \textit{Slope Inclination:} Terrain slopes affect sensor performance, particularly the field of view for the cameras. Steeper slopes can cause the sensors to either point too far downwards, focusing on the grass, or upwards, potentially capturing images of the sky, which can be challenging for effective localization.
    
    \item \textit{Long-term Stability:} This criterion assesses how stable the environment remains over time. Changes such as vegetation growth or alterations in the surroundings can impact the reliability of visual localization systems by altering the availability of visual landmarks.
    
    \item \textit{Vegetation:} The presence of trees, shrubs, and other forms of vegetation provide both obstacles and landmarks for the robot. High vegetation complexity can enhance localization by offering distinctive features but can also introduce occlusions and navigational difficulties.
    
    \item \textit{Brightness:} The level of ambient light directly impacts the effectiveness of visual localization systems. Bright conditions offer consistent data, while shaded or low-light areas can hinder perception. Dynamic changes, such as shifting brightness in partly cloudy weather, add further challenges to maintaining accurate localization.
\end{itemize}


\begin{table*}[!hb]
    \centering
    \caption{Characteristics of recording locations, including size, terrain, and environmental features.}
    \fontsize{8pt}{8pt}\selectfont
    \resizebox{\textwidth}{!}{%
        \begin{tabular*}{1.1\textwidth}{@{\extracolsep{\fill}} lcccccc} 
         \toprule
            \multirow{ 2}{*}{\textbf{Location}} & \multirow{ 2}{*}{\textbf{Size [sqm]}} & \textbf{Shape} & \textbf{Slope} & \textbf{Long-Term} & \multirow{ 2}{*}{\textbf{Vegetation}} & \multirow{ 2}{*}{\textbf{Brightness}} \\
            ~ & ~ & \textbf{Complexity} & \textbf{Inclination} & \textbf{Stability} & ~ & ~ \\
            \toprule
            Park & 250 & High & Medium & Medium & High & Medium \\ 
            Campus Small & 75 & Medium & None & High & Low & Low \\ 
            Campus Large & 500 & Low & None & High & Low & High \\ 
            Garden Small & 100 & Medium & None & Medium & Medium & Low \\ 
            Garden Large & 250 & High & Low & Low & High & High \\
            \bottomrule
        \end{tabular*}%
    } 
    \label{table:locations}
    \centering
    \vspace{2em}
    \caption{Recording overview across locations and environmental conditions.}
    \fontsize{8pt}{8pt}\selectfont
    \resizebox{\textwidth}{!}{%
        \begin{tabular*}{1.1\textwidth}{@{\extracolsep{\fill}} lccccccccc} 
            \toprule
            \textbf{Location} & \textbf{Date} & \textbf{Distance [m]} & \textbf{Duration [s]} & \textbf{Windy} & \textbf{Sunny} & \textbf{Cloudy} & \textbf{Dusk} & \textbf{Night} & \textbf{+ Light}\\
            \toprule

            \multirow{7}{*}{Park} & 2023-07-31 - 11:28:49 & 164,24 & 422,71 & ~ & ~ & x & ~ & ~ & ~ \\ 
            & 2023-11-07 - 16:07:01 & 170,47 & 482,32 & x & ~ & x & ~ & ~ & ~ \\ 
            & 2024-04-14 - 14:56:43 & 171,46 & 446,40 & ~ & x & x & ~ & ~ & ~ \\ 
            & 2024-05-08 - 12:00:23 & 183,00 & 466,35 & ~ & ~ & x & ~ & ~ & ~ \\ 
            & 2024-05-13 - 21:09:16 & 182,67 & 463,27 & ~ & ~ & ~ & x & ~ & ~ \\ 
            & 2024-05-13 - 21:28:35 & 179,75 & 457,37 & ~ & ~ & ~ & ~ & x & ~ \\ 
            & 2024-05-24 - 21:53:02 & 172,13 & 443,30 & ~ & ~ & ~ & ~ & ~ & x \\ \midrule
            \multirow{8}{*}{Campus Small} & 2023-09-11 - 09:39:05 & 155,27 & 403,77 & ~ & x & ~ & ~ & ~ & ~ \\ 
            & 2023-11-23 - 15:41:16 & 156,16 & 405,85 & ~ & ~ & x & ~ & ~ & ~ \\ 
            & 2024-02-19 - 15:11:40 & 149,77 & 379,42 & ~ & ~ & x & ~ & ~ & ~ \\ 
            & 2024-04-14 - 14:34:47 & 151,87 & 387,61 & ~ & x & x & ~ & ~ & ~ \\ 
            & 2024-05-07 - 12:12:13 & 156,28 & 400,38 & ~ & ~ & x & ~ & ~ & ~ \\ 
            & 2024-05-08 - 20:55:29 & 157,31 & 399,63 & ~ & ~ & ~ & x & ~ & ~ \\ 
            & 2024-05-08 - 21:18:04 & 155,07 & 397,84 & ~ & ~ & ~ & ~ & x & ~ \\ 
            & 2024-05-24 - 22:09:50 & 153,60 & 395,17 & ~ & ~ & ~ & ~ & ~ & x \\ \midrule
            \multirow{8}{*}{Campus Large} & 2023-07-20 - 13:50:02 & 384,92 & 1035,23 & x & ~ & x & ~ & ~ & ~ \\ 
            & 2023-11-07 - 15:27:56 & 381,24 & 978,09 & x & ~ & x & ~ & ~ & ~ \\ 
            & 2024-01-27 - 10:19:38 & 289,71 & 740,68 & ~ & x & ~ & ~ & ~ & ~ \\ 
            & 2024-04-14 - 14:01:26 & 368,39 & 928,93 & ~ & x & x & ~ & ~ & ~ \\ 
            & 2024-09-24 - 19:36:44 & 277,29 & 707,99 & ~ & ~ & x & x & ~ & ~ \\ 
            & 2024-09-24 - 19:55:55 & 277,81 & 692,71 & ~ & ~ & ~ & ~ & x & ~ \\ 
            & 2024-09-24 - 20:09:32 & 274,46 & 695,15 & ~ & ~ & ~ & ~ & ~ & x \\ 
            & 2024-09-25 - 14:27:59 & 277,19 & 701,29 & x & ~ & x & ~ & ~ & ~ \\ \midrule
            \multirow{8}{*}{Garden Small} & 2023-08-18 - 14:27:36 & 116,36 & 323,18 & x & x & ~ & ~ & ~ & ~ \\ 
            & 2023-09-15 - 16:37:27 & 116,34 & 320,92 & ~ & ~ & x & ~ & ~ & ~ \\ 
            & 2024-01-13 - 09:17:46 & 113,30 & 330,22 & ~ & ~ & x & ~ & ~ & ~ \\ 
            & 2024-04-11 - 16:52:39 & 124,79 & 338,36 & ~ & x & ~ & ~ & ~ & ~ \\ 
            & 2024-05-29 - 17:24:28 & 121,35 & 310,83 & ~ & ~ & x & ~ & ~ & ~ \\ 
            & 2024-05-29 - 21:14:15 & 124,21 & 312,69 & ~ & ~ & ~ & x & ~ & ~ \\ 
            & 2024-05-29 - 21:35:57 & 122,68 & 315,26 & ~ & ~ & ~ & ~ & x & ~ \\ 
            & 2024-05-29 - 21:47:30 & 120,04 & 307,88 & ~ & ~ & ~ & ~ & ~ & x \\ \midrule
            \multirow{8}{*}{Garden Large} & 2023-08-18 - 14:54:17 & 167,66 & 452,58 & x & x & ~ & ~ & ~ & ~ \\ 
            & 2023-12-21 - 11:23:58 & 170,41 & 446,20 & ~ & ~ & x & ~ & ~ & ~ \\ 
            & 2024-01-13 - 10:02:43 & 162,28 & 415,65 & ~ & ~ & x & ~ & ~ & ~ \\ 
            & 2024-04-11 - 17:06:47 & 164,99 & 431,86 & ~ & x & ~ & ~ & ~ & ~ \\ 
            & 2024-05-29 - 17:13:26 & 150,31 & 377,22 & ~ & ~ & x & ~ & ~ & ~ \\ 
            & 2024-05-29 - 21:20:14 & 151,83 & 385,43 & ~ & ~ & ~ & x & ~ & ~ \\ 
            & 2024-05-30 - 21:50:12 & 150,53 & 380,96 & ~ & ~ & ~ & ~ & x & ~ \\ 
            & 2024-05-30 - 22:03:19 & 151,80 & 383,83 & ~ & ~ & ~ & ~ & ~ & x \\

            \bottomrule
        \end{tabular*}%
    } 
    \label{table:recordings}
\end{table*}

Figure~\ref{fig:environments} illustrates the diverse outdoor settings used in the dataset, chosen to represent a range of environmental challenges for SLAM evaluation. The \textit{Park}, \textit{Campus Small}, and \textit{Campus Large} are located on the Esslingen University campus, providing a variety of terrain and structural features. In contrast, the \textit{Garden Large} and \textit{Garden Small} settings, situated in a private garden, add complexity with natural vegetation and varied layouts. Together, these locations allow for comprehensive testing across different brightness levels, vegetation densities, and spatial characteristics, summarized in Table~\ref{table:locations}.
The \textit{Park} presents moderate slopes and dense vegetation, with natural obstacles such as trees and bushes that aid navigation but can also obstruct movement. The ambient brightness level in this setting adds to the challenge of operating in a dynamic, natural environment. \textit{Campus Small} is a compact lawn adjacent to a building, featuring minimal vegetation and a reflective glass surface that introduces challenges for visual localization. Shadows cast by nearby structures create a low-brightness environment that simplifies testing in controlled conditions. \textit{Campus Large}, a rooftop space with uneven terrain and minimal vegetation, offers an expansive area with a horizon-like view, limiting distinctive visual cues. The stable brightness and open space of this location make it a useful testbed for navigation in settings with few visual landmarks.
In the private garden, \textit{Garden Large} features a traditional garden layout with dense vegetation, low incline, and ample light. This setting provides rich visual diversity, making it ideal for challenging navigation and path-planning tasks. In contrast, \textit{Garden Small}, shaded by surrounding buildings and with a uniform wall on one side, limits visual cues and presents moderate vegetation density with reduced brightness. This setup increases the difficulty of maintaining consistent tracking, highlighting the challenges posed by more confined, less varied environments.

\begin{figure*}[t!]
  \centering
  \footnotesize
  \setlength{\tabcolsep}{0.05cm}
  {\renewcommand{\arraystretch}{1}
    \begin{tabular}{p{4.4cm} p{4.4cm} p{4.4cm} p{4.4cm}}
        \includegraphics[width=4.4cm]{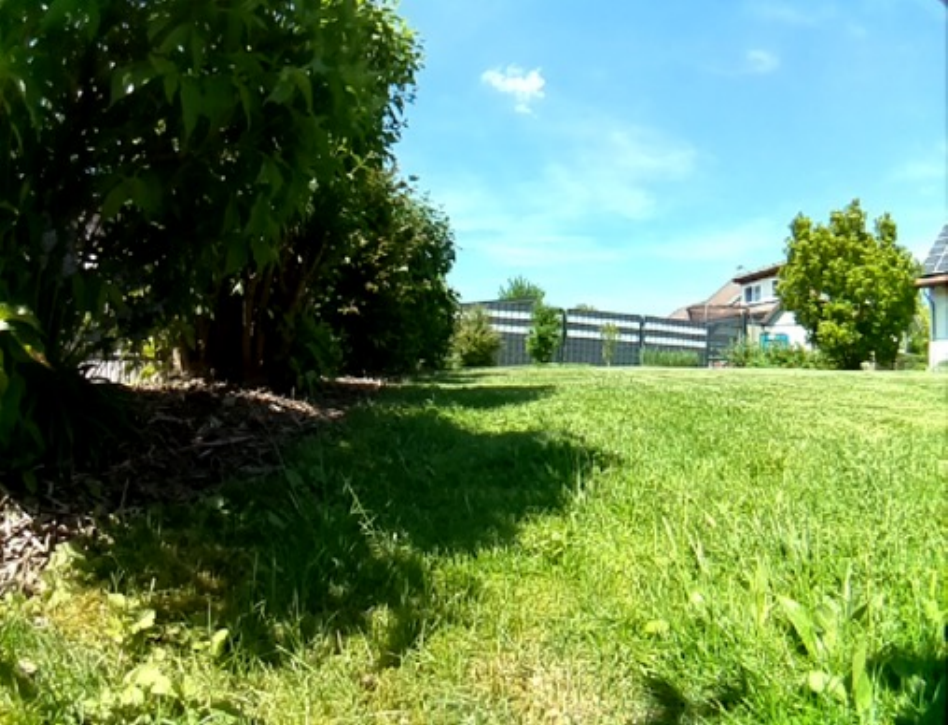} &
        \includegraphics[width=4.4cm]{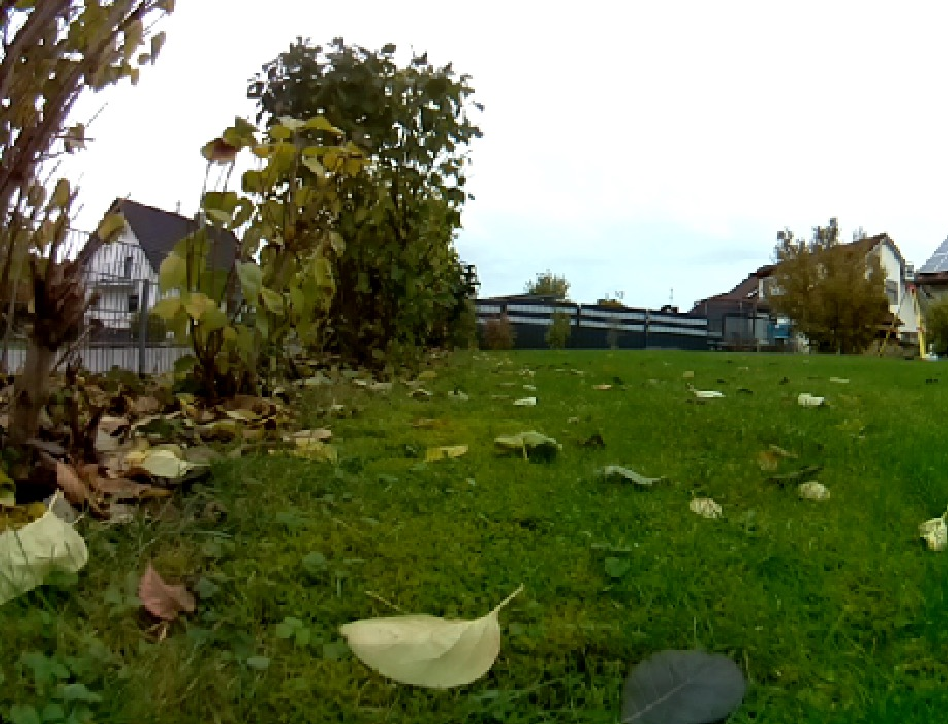} &
        \includegraphics[width=4.4cm]{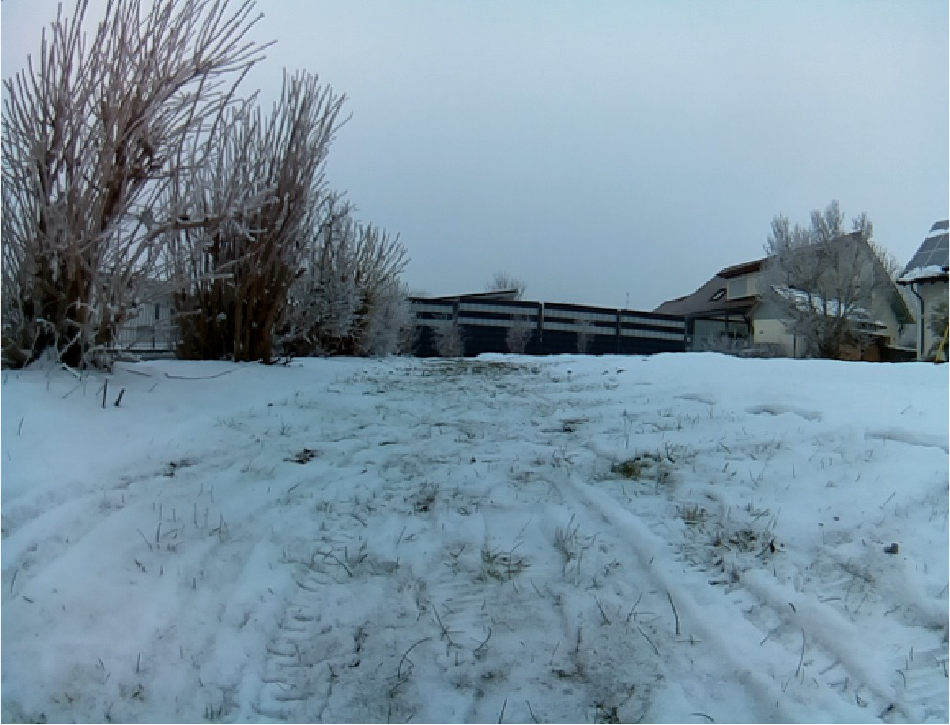} &
        \includegraphics[width=4.4cm]{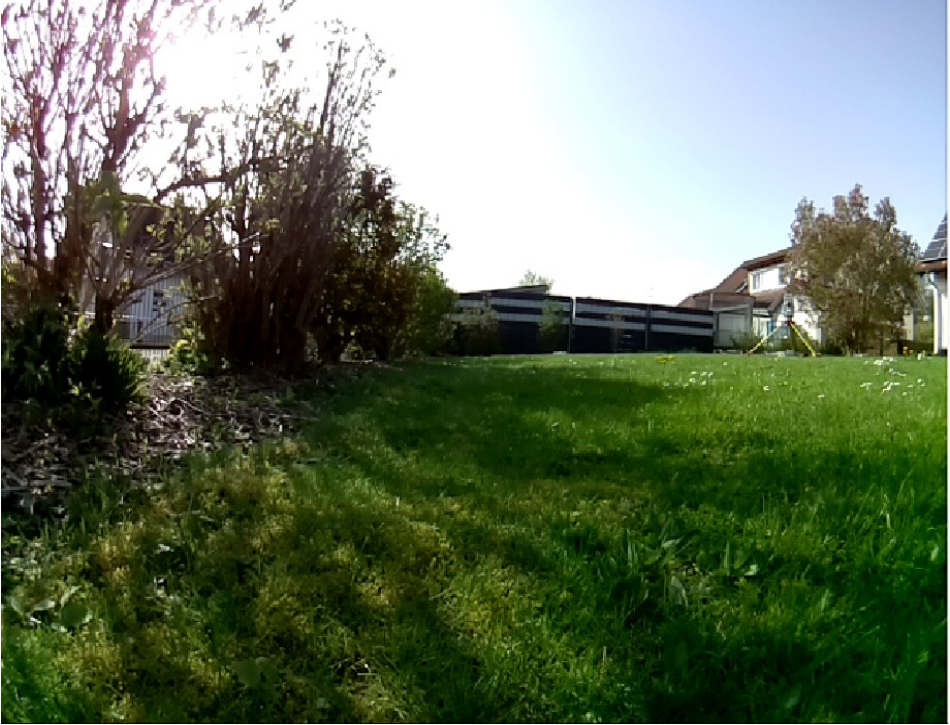} \\
        \multicolumn{1}{c}{(a) Summer} &
        \multicolumn{1}{c}{(b) Autumn} &
        \multicolumn{1}{c}{(c) Winter} &
        \multicolumn{1}{c}{(d) Spring} \\[0.25em]

        \includegraphics[width=4.4cm]{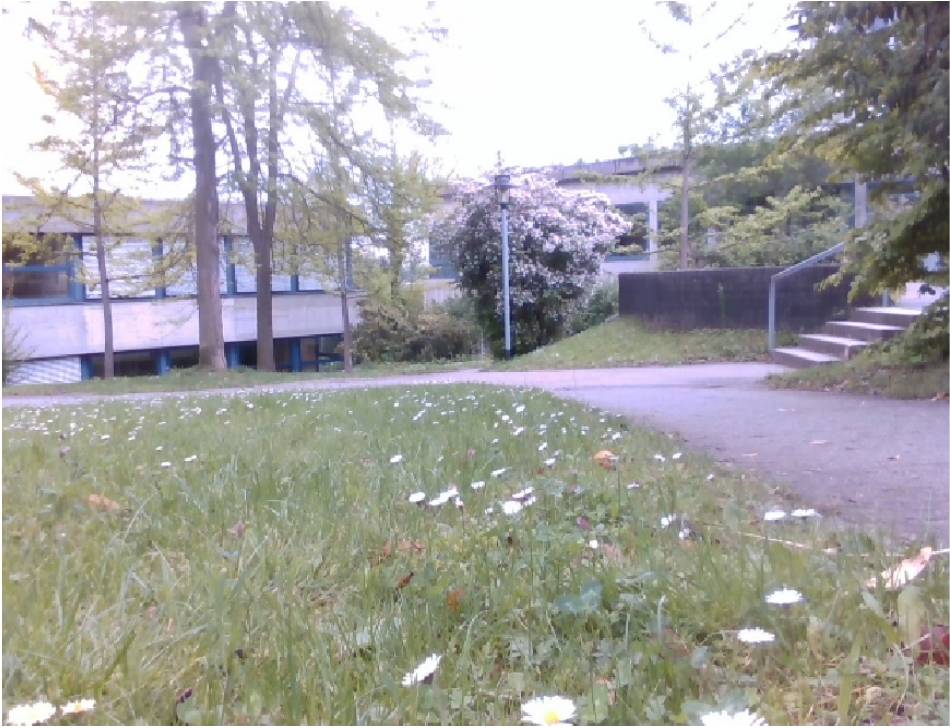} &
        \includegraphics[width=4.4cm]{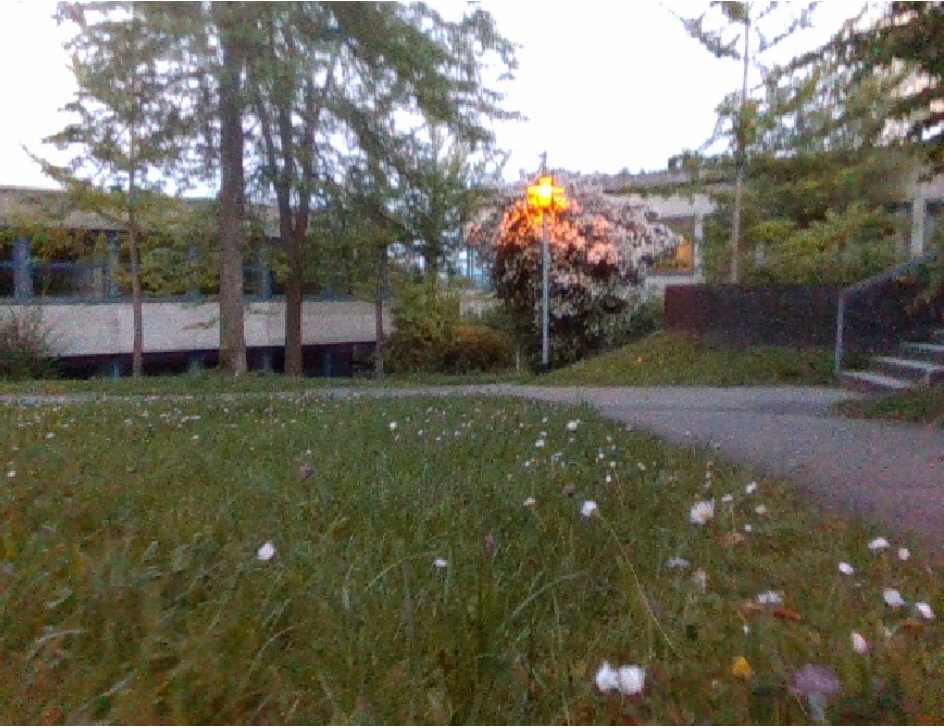} &
        \includegraphics[width=4.4cm]{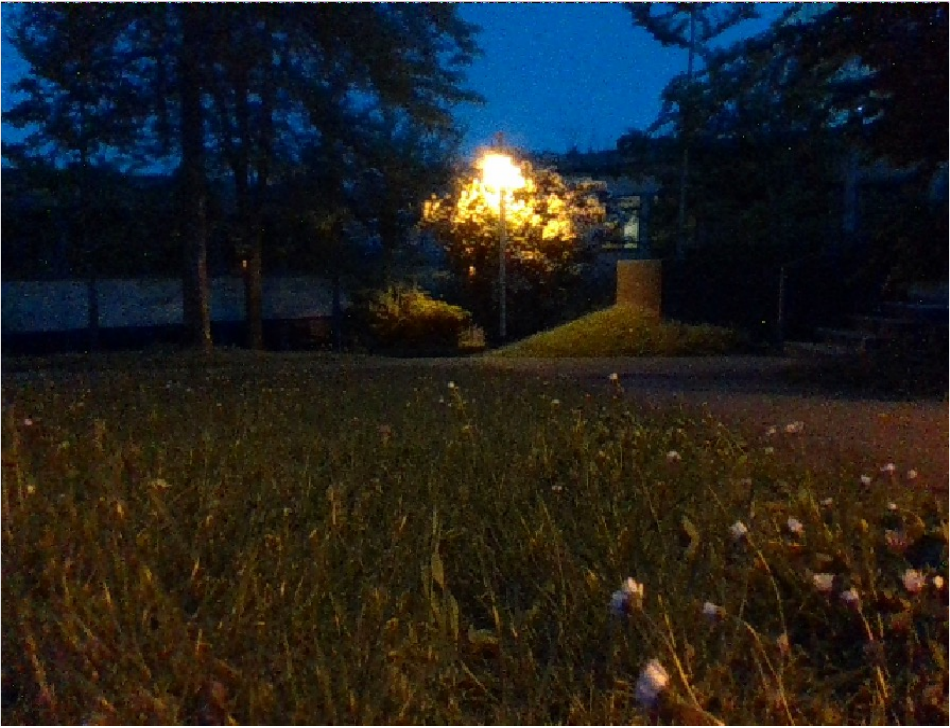} &
        \includegraphics[width=4.4cm]{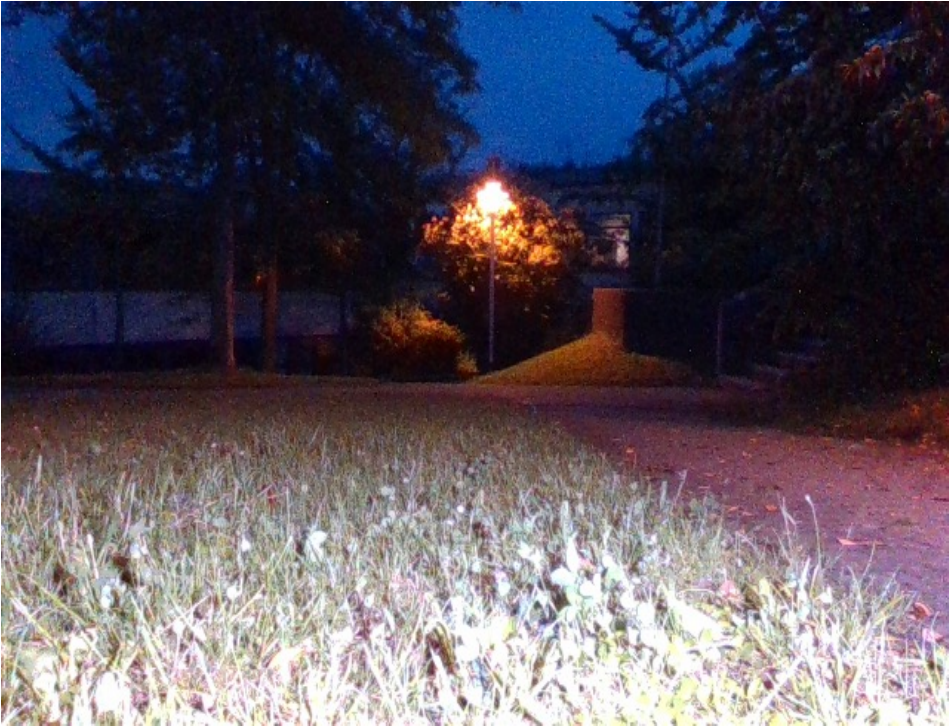} \\
        \multicolumn{1}{c}{(e) Day} &
        \multicolumn{1}{c}{(f) Dusk} &
        \multicolumn{1}{c}{(g) Night} &
        \multicolumn{1}{c}{(h) Night + Light} \\
    \end{tabular}}
    \caption{Preview of our dataset capturing diverse environmental conditions and seasonal variations across (a) Summer, (b) Autumn, (c) Winter, and (d) Spring, along with lighting conditions of (e) Day, (f) Dusk, (g) Night, and (h) Night with external lighting. These settings illustrate the range of challenges for evaluating SLAM algorithms under different natural conditions.}
    \label{fig:dataset_preview}
\end{figure*}

\subsection{Recording Overview}
The dataset captures each location’s variations across seasons and weather conditions, focusing on how factors like vegetation, brightness, and lighting impact the robot's performance. Table~\ref{table:recordings} summarizes the recordings and their associated environmental conditions, while Figure~\ref{fig:dataset_preview} illustrates seasonal appearances and diverse lighting conditions within the dataset. Each location was recorded in a variety of weather conditions, from sunny to cloudy to windy, and at different times of day, including dusk and night, to evaluate the effects of sunlight and low-light conditions on sensor performance.

To ensure consistency, we employed the \textit{Perimeter} scenario, where the robot completed three full rounds along the edge of the lawn area. This scenario results in successive rounds with slight trajectory differences and partial overlaps, offering realistic spatial coverage variation. This approach allows us to analyze how environmental changes over time affect localization accuracy along the same path. It also enables loop closing opportunities for maintaining accurate long-term localization and supports long-term mapping by capturing the same environment repeatedly under changing conditions.

\subsection{Dataset Organisation}


\highlight{The dataset includes recordings from multiple locations and dates (see Table \ref{table:recordings}). For each combination of \texttt{<location>} and \texttt{<date>}, the data is organized into folders following the hierarchy illustrated in Figure \ref{fig:dirtree}. Inside each \texttt{<date>} folder, we provide separate directories for the D435i, the T265, and the PiCam, as well as a folder for VN100 IMU data and a file for the 3~DOF ground truth trajectory (\texttt{groundtruth.txt}).

All RGB and depth images for the D435i are named by their Unix timestamp (\texttt{<timestamp>}) and saved as a PNG file. We include \texttt{rgb.txt} and \texttt{depth.txt} files following the format proposed by~\cite{sturm2012tumrgbd}, where each line contains a timestamp and an image file path to facilitate easy parsing by common SLAM frameworks. IMU data from the D435i is stored in \texttt{imu.txt}, containing the 6~DoF measurements (timestamped linear acceleration and angular velocity).
The T265 is organized analogous to the D435i. Two image directories (\texttt{cam\_left} and \texttt{cam\_right}) hold stereo PNG files, with matching \texttt{cam\_left.txt} and \texttt{cam\_right.txt} for timestamp-to-file mapping. The T265’s IMU measurements are stored in \texttt{imu.txt}, with the same format as the D435i.
Similarly, the PiCam data is in the \texttt{picam} directory, with RGB images named by their \texttt{<timestamp>}, plus a file \texttt{rgb.txt} for time-indexed image paths.
The \texttt{vn100} folder contains a single \texttt{imu.txt} file that logs the timestamped 9~DoF (accelerometer, gyroscope, magnetometer) readings of the VN100 sensor. 
All calibration files are stored in the \texttt{calibration} directory, with one YAML file per sensor (e.g., \texttt{realsense\_D435i.yaml}). These files contain the intrinsic parameters for each sensor; for the D435i and T265, they also include the extrinsic calibration to the internal IMU. In addition, all cameras are calibrated with respect to both the VN100 and the prism.

In addition to the raw file-based data, we provide a Python script to convert each folder’s contents into rosbags. Table~\ref{table:rosbag} outlines the ROS topic structure, message types, and approximate publish rates for each sensor. This enables users to integrate smoothly with ROS-based SLAM and perception frameworks.}

\begin{figure}
    \fontsize{8pt}{8pt}\selectfont
    \dirtree{%
    .1 /ROVER.
    .2 <location>.
    .3 <date>.
    .4 realsense\_D435i.
    .5 rgb.
    .6 <timestamp>.png.
    .5 depth.
    .6 <timestamp>.png.
    .5 rgb.txt.
    .5 depth.txt.
    .5 imu.
    .6 imu.txt.
    .4 realsense\_T265.
    .5 cam\_left.
    .6 <timestamp>.png.
    .5 cam\_right.
    .6 <timestamp>.png.
    .5 cam\_left.txt.
    .5 cam\_right.txt.
    .5 imu.
    .6 imu.txt.
    .4 picam.
    .5 rgb.
    .6 <timestamp>.png.
    .5 rgb.txt.
    .4 vn100.
    .5 imu.txt.
    .4 groundtruth.txt.
    .2 calibration.
    .3 realsense\_D435i.yaml.
    .3 realsense\_T265.yaml.
    .3 picam.yaml.
    .3 vn100.yaml.
    }
    
    \caption{\highlight{Folder structure of the ROVER dataset.}}
    \label{fig:dirtree}
\end{figure}

\begin{table*}[!t]
    \centering
    \caption{\highlight{Rosbag overview: topics, message types, and publish rates.}}
        \begin{tabular*}{\textwidth}{@{\extracolsep{\fill}} llll} 
         \toprule
            \textbf{Sensor} & \textbf{ROS Topic} & \textbf{ROS Message Type} & \textbf{Publish Rate (hz)} \\
            \toprule
            \multirow{3}{*}{Intel RealSense D435i} & \texttt{/d435i/rgb\_image} & \texttt{sensor\_msgs/Image} & 30 \\
             & \texttt{/d435i/depth\_image} & \texttt{sensor\_msgs/Image} & 30 \\ 
             & \texttt{/d435i/imu} & \texttt{sensor\_msgs/Imu} & 300 \\ 
            \midrule
            \multirow{3}{*}{Intel RealSense T265} & \texttt{/t265/image\_left} & \texttt{sensor\_msgs/Image} & 30 \\
             & \texttt{/t265/image\_left} & \texttt{sensor\_msgs/Image} & 30 \\ 
             & \texttt{/t265/imu} & \texttt{sensor\_msgs/Imu} & 265 \\
            \midrule
            PiCam & \texttt{/pi\_cam/rgb\_image} & \texttt{sensor\_msgs/Image} & 30 \\
            \midrule
            VN100 & \texttt{/vn100/imu} & \texttt{sensor\_msgs/Imu} & 65 \\
            \bottomrule
        \end{tabular*}%
    \label{table:rosbag}
\end{table*}

\section{Visual SLAM Benchmark}
\label{sec:benchmark}

Evaluating SLAM algorithms across diverse conditions is essential to understanding their robustness and adaptability in real-world applications. This section presents a benchmark framework for assessing various SLAM approaches, from traditional methods to deep learning-based techniques, under a wide range of environmental and sensor configurations. The following subsections outline the selected algorithms, describe the evaluation methodology, and analyze results to highlight accuracy, resilience, and the specific challenges the algorithms face in different settings. 

\subsection{Visual SLAM Methods}
\label{sec:slam_methods}
We explore a variety of visual and visual-inertial SLAM methods, employing both traditional and deep learning-based approaches, to address different aspects of visual localization and mapping. The methods differ in their design, ranging from feature-based and semi-direct techniques to deep learning methods that utilize dense direct approaches with optical flow estimation. Additionally, some methods incorporate loop closing for long-term consistency, while others do not. The approaches also vary in estimation techniques, utilizing either filter-based or optimization-based backends to support different sensor modalities, such as monocular, stereo, and RGBD, with or without inertial data integration. Table~\ref{table:slam_methods} provides a detailed overview of the capabilities and configurations of the SLAM methods considered in this study.

\begin{table*}[htb]
    \centering
    \caption{Overview of SLAM methods, supported sensor configurations, and algorithmic components.}
    \fontsize{8pt}{8pt}\selectfont
    \setlength\extrarowheight{1mm}
    \resizebox{\textwidth}{!}{%
        \begin{tabular*}{1.1\textwidth}{@{\extracolsep{\fill}} lcccccccccc} 
            \toprule
            \multirow{ 2}{*}{\textbf{SLAM}} & \multirow{ 2}{*}{\textbf{Mono}} & \textbf{Mono-} & \multirow{ 2}{*}{\textbf{Stereo}} & \textbf{Stereo-} & \multirow{ 2}{*}{\textbf{RGBD}} & \textbf{RGBD-} & \multirow{ 2}{*}{\textbf{Frontend}} & \multirow{ 2}{*}{\textbf{Estimation}} & \multirow{ 2}{*}{\textbf{Backend}} & \textbf{Loop} \\
            ~ & ~ & \textbf{inertial} & ~ & \textbf{inertial} & ~ & \textbf{inertial} & ~ & ~ & ~ & \textbf{Closing} \\
            \toprule

            OpenVINS \cite{geneva2020openvins}      & ~ & x & ~ & x & ~ & ~ & Sparse (Feature) & MSCKF & ~ & ~ \\ 
            VINS-Fusion \cite{qin2019vins-fusion}   & ~ & x & x & x & ~ & ~ & Sparse (Feature) & Local BA & PGO & DBoW2 \\ 
            ORB-SLAM3 \cite{campos2021orb-slam3}    & x & x & x & x & x & x & Sparse (Feature) & Local BA & Global BA \& PGO & DBoW2 \\ 
            SVO Pro \cite{forster2017svo}       & x & x & x & x & ~ & ~ & Sparse (Semi-Direct) & Local BA & Global BA / PGO & DBoW2 \\ 
            DROID-SLAM \cite{teed2021droid-slam}    & x & ~ & x & ~ & x & ~ & Dense (Optical Flow) & Local BA & Global BA \& PGO & ~ \\ 
            DPVO \cite{teed2023dpvo}                & x & ~ & ~ & ~ & ~ & ~ & Dense (Optical Flow) & Local BA & ~ & ~ \\ 
            DPV-SLAM \cite{lipson2024dpv-slam}      & x & ~ & ~ & ~ & ~ & ~ & Dense (Optical Flow) & Local BA & Global BA \& PGO & DBoW2 \\

            \bottomrule
        \end{tabular*}%
    } 
    \label{table:slam_methods}
\end{table*}

Traditional SLAM methods, such as OpenVINS~\cite{geneva2020openvins}, VINS-Fusion~\cite{qin2019vins-fusion}, ORB-SLAM3~\cite{campos2021orb-slam3}, and SVO Pro~\cite{forster2017svo}, provide robust baselines using well-established algorithms. These approaches often rely on Bundle Adjustment (BA)~\cite{triggs2000bundle} to refine camera poses and 3D point estimates and use Pose Graph Optimization~(PGO)~\cite{grisetti2010pgo} for improving trajectory consistency, especially in long-term mapping. Some also incorporate the Multi-State Constraint Kalman Filter~(MSCKF)~\cite{mourikis2007msckf}, a filter-based approach that fuses visual and inertial data to estimate camera motion. For loop closing, methods like ORB-SLAM3 use Dynamic Bag-of-Words (DBoW2)~\cite{galvez2012bow}, an algorithm that employs a visual vocabulary to recognize previously visited locations based on visual features.
Deep learning-based methods, such as DROID-SLAM~\cite{teed2021droid-slam}, DPVO~\cite{teed2023dpvo}, and DPV-SLAM~\cite{lipson2024dpv-slam}, in contrast, introduce data-driven techniques for more adaptive mapping, leveraging dense optical flow for tracking and estimation. In comparison, CL-SLAM~\cite{vodisch2022continual} and COVIO~\cite{vodisch2023covio} leverage continual learning to adapt during deployment in an online manner.

\subsection{Evaluation Methodology}

To account for the variability in SLAM algorithm performance, each dataset sequence is tested across five trials. A trajectory filtering process is then applied to exclude failed runs, ensuring a reliable evaluation of the results.
The filtering criteria are designed to preserve data integrity by focusing on trajectories that adequately reflect SLAM system performance. Specifically, a trajectory is considered valid if it has pose estimates covering at least 80\% of the scenario duration, ensuring sufficient coverage and consistency. Additionally, the trajectory must provide at least one pose per second to maintain adequate temporal resolution for precise tracking. Any runs failing to meet these criteria are excluded from further analysis. To quantify the effectiveness of various SLAM algorithms, we employ the concept of success rate, as initially introduced in~\cite{wang2020tartanair}. This metric is adapted to quantify the ratio of valid trajectories generated to the total number of trials. By indicating how frequently trajectories satisfy filtering criteria, the success rate emphasizes an algorithm's robustness across multiple trials and diverse dataset conditions.

Once the valid trajectories are identified, alignment is performed to accurately compare the estimated and ground truth trajectories. Since the trajectories may exist in different coordinate systems, we employ Umeyama's method~\cite{umeyama1991alignment} to find the optimal transformation that maps the estimated trajectory onto the ground truth, yielding a least-squares fit. Following Zhang and Scaramuzza~\cite{zhang2018tutorial}, SLAM methods rely solely on monocular systems, the trajectories are scaled relative to the ground truth using a SIM(3) transformation to account for scale ambiguity. In contrast, for other modalities such as stereo, RGBD, or those incorporating inertial data, where the scale is observable, an SE(3) transformation (including only rotation and translation) is applied. This alignment step ensures that the subsequent evaluation measures the true deviation from the ground truth.

Finally, we calculate the root mean squared error (RMSE) of the Absolute Trajectory Error (ATE) \highlight{and the RMSE of the Relative Pose Error (RPE) per meter as the primary metrics to quantify the global and local accuracy of the SLAM system,} following the methodology outlined in~\cite{sturm2012tumrgbd}. To ensure temporal consistency, linear interpolation is applied to synchronize the timestamps between the estimated and ground truth data points before computing the RMSE. The RMSE ATE \highlight{and RPE} are calculated using the open-source Evo toolbox~\cite{grupp2017evo}. \highlight{In addition, we calculate the success rate (SR), which serves as a key indicator of a method's robustness by reflecting the ratio of valid trajectories to the total number of trials for each SLAM configuration.} All experiments were conducted on an AMD Ryzen 9 7950X3D with 64~GB of RAM and an NVIDIA RTX A6000 within a Docker container based on Ubuntu~20.04 and ROS Noetic. We utilized open-source implementations of each SLAM algorithm and applied pretrained weights for DROID-SLAM, DPVO, and DPV-SLAM to ensure reproducibility and consistency in the evaluation.

\subsection{Experiments and Results}

We present the results of the experimental evaluation of the SLAM algorithms across a range of sensor configurations and environmental conditions. Each SLAM algorithm was tested using all supported sensor configurations in Table~\ref {table:slam_methods}, combining different cameras and IMUs on all of the 39 sequences recorded. For each SLAM configuration, we compute the ATE \highlight{and RPE} for every location (averaged across the four seasons) \highlight{and then determine the overall mean ATE (mATE), mean RPE (mRPE), and SR across all locations.}
To highlight the performance of each SLAM method, we select a representative sensor configuration for each algorithm, specifically the configuration that achieves the lowest \highlight{mATE}. For these selected configurations, we further analyze their performance under different environmental conditions, i.e., varying seasons and lighting scenarios, to examine how these factors influence SLAM accuracy.

\begin{table*}[!ht]
\centering
\caption{\highlight{mATE [m], mRPE [m] and SR [\%] for mono SLAM configurations. The best results for mATE and mRPE are shown in bold.}}
\begin{tabular}{l ccc ccc ccc}
    \toprule
    \textbf{SLAM} 
    & \multicolumn{3}{c}{\textbf{PiCam}} 
    & \multicolumn{3}{c}{\textbf{D435i}} 
    & \multicolumn{3}{c}{\textbf{T265}} \\
    \cmidrule(lr){2-4} \cmidrule(lr){5-7} \cmidrule(lr){8-10}
     & mATE & mRPE & SR  
     & mATE & mRPE & SR 
     & mATE & mRPE & SR \\
    \midrule
    ORB-SLAM3  
      & 7.82 & 4.05 & 78.95  
      & \textbf{3.60} & \textbf{1.34} & 26.32  
      & \textbf{4.05} & \textbf{0.50} & 94.74 \\
    SVO Pro    
      & \textbf{7.31} & 2.94 & 84.21 
      & 7.94 & 3.29 & 100.00
      & 6.78 & 5.56 & 94.74 \\
    DROID-SLAM 
      & 7.76 & 4,24 & 94.74
      & 5.63 & 1,73 & 94.74
      & 77.28 & 4.75 & 26.32 \\
    DPVO       
      & 7.79 & 3.43 & 94.74
      & 6.47 & 2,15 & 100.00
      & 8.22 & 5.48 & 100.00 \\
    DPV-SLAM   
      & 7.77 & \textbf{1.63} & 94.74
      & 6.24 & 1.84 & 100.00
      & 8.21 & 5.70 & 100.00 \\
    \bottomrule
\end{tabular}
\label{table:mono_results}
\end{table*}

\begin{table*}[!ht]
\centering
\caption{\highlight{Mono-inertial SLAM results. An 'x' indicates that no valid trajectories were generated by the SLAM configuration.}}
\resizebox{\textwidth}{!}{%
\begin{tabular}{l ccc ccc ccc ccc ccc}
\toprule
\textbf{SLAM} 
& \multicolumn{3}{c}{\textbf{PiCam - external}} 
& \multicolumn{3}{c}{\textbf{D435i - internal}} 
& \multicolumn{3}{c}{\textbf{D435i - external}} 
& \multicolumn{3}{c}{\textbf{T265 - internal}} 
& \multicolumn{3}{c}{\textbf{T265 - external}} \\
\cmidrule(lr){2-4}
\cmidrule(lr){5-7}
\cmidrule(lr){8-10}
\cmidrule(lr){11-13}
\cmidrule(lr){14-16}
& mATE & mRPE & SR
& mATE & mRPE & SR
& mATE & mRPE & SR
& mATE & mRPE & SR
& mATE & mRPE & SR \\
\midrule
OpenVINS
& \textbf{1.74} & \textbf{0.17} & 78.95
& \textbf{4.14} & \textbf{0.19} & 68.42
& 23.53 & \textbf{0.24} & 21.05
& \textbf{1.49} & \textbf{0.16} & 94.74
& \textbf{1.82} & \textbf{0.15} & 100.00 \\

VINS-Fusion
& x & x & 0.00
& \textbf{4.14} & 0.98 & 73.68
& x & x & 0.00
& 8.16 & 0.83 & 21.05
& 4.89 & 0.32 & 21.05 \\

ORB-SLAM3
& x & x & 0.00
& x & x & 0.00
& \textbf{15.32} & 1.54 & 10.53
& 19.15 & 1.46 & 36.84
& 2.69 & 0.38 & 21.05 \\

SVO Pro
& 3.29 & 0.18 & 57.89
& x & x & 0.00
& 24.19 & 0.26 & 15.79
& 2.62 & 0.17 & 63.16
& 15.58 & 0.19 & 42.11 \\

\bottomrule
\end{tabular}
}
\label{table:Mono-inertial_results}
\end{table*}

\begin{table*}[!ht]
\centering
\caption{\highlight{Stereo and stereo-inertial SLAM results. 'n/a' indicates that the sensor configuration is not supported by the SLAM.}}
\begin{tabular}{l ccc ccc ccc}
    \toprule
    \multirow{2}{*}{\textbf{SLAM}} 
      & \multicolumn{3}{c}{\textbf{Stereo}} 
      & \multicolumn{3}{c}{\textbf{Stereo-inertial (internal)}} 
      & \multicolumn{3}{c}{\textbf{Stereo-inertial (external)}} \\
    \cmidrule(lr){2-4} \cmidrule(lr){5-7} \cmidrule(lr){8-10}
    & mATE & mRPE & SR
    & mATE & mRPE & SR
    & mATE & mRPE & SR \\
    \midrule
    
    OpenVINS    
    & n/a & n/a & n/a  
    & \textbf{1.43} & 0.23 & 94.74 
    & 1.74 & 0.27 & 100.00 \\
    
    VINS-Fusion 
    & x & x & 0.00
    & 3.13 & 0.57 & 73.68
    & 3.26 & 0.28 & 5.26 \\
    
    ORB-SLAM3  
    & \textbf{2.41} & \textbf{0.35} & 84.21
    & 7.51 & 0.48 & 47.37
    & \textbf{0.71} & \textbf{0.14} & 42.11 \\
    
    SVO Pro     
    & 49.89 & 0.38 & 73.68
    & 4.21 & \textbf{0.17} & 68.42
    & 1.70 & 0.16 & 100.00 \\
    
    DROID-SLAM  
    & 77.28 & 0.76 & 26.32
    & n/a & n/a & n/a
    & n/a & n/a & n/a \\
    
    \bottomrule
\end{tabular}

\label{table:stereo_results}
\end{table*}


\begin{table*}[!ht]
\centering
\caption{\highlight{RGBD and RGBD-inertial SLAM results.}}
\begin{tabular}{l ccc ccc ccc}
    \toprule
    \multirow{2}{*}{\textbf{SLAM}} 
      & \multicolumn{3}{c}{\textbf{RGBD}} 
      & \multicolumn{3}{c}{\textbf{RGBD-inertial (internal)}} 
      & \multicolumn{3}{c}{\textbf{RGBD-inertial (external)}} \\
    \cmidrule(lr){2-4} \cmidrule(lr){5-7} \cmidrule(lr){8-10}
    & mATE & mRPE & SR
    & mATE & mRPE & SR
    & mATE & mRPE & SR \\
    \midrule
    
    ORB-SLAM3  
    & \textbf{1.26} & \textbf{0.17} & 94.74 
    & \textbf{5.94} & \textbf{0.18} & 73.68 
    & \textbf{8.35} & \textbf{0.33} & 78.95 \\
    
    DROID-SLAM 
    & 2.32 & 0.21 & 94.74
    & n/a & n/a & n/a
    & n/a & n/a & n/a \\
    
    \bottomrule
\end{tabular}
\label{table:RGBD_results}
\end{table*}

The evaluation of SLAM configurations reveals several important trends across different sensor setups and algorithms. 
In the mono configurations (see Table~\ref{table:mono_results}), there is a general trend of higher \highlight{mATE and mRPE} values across all camera types. \highlight{Notably, ORB-SLAM3 using the D435i camera achieves the lowest mATE, which indicates the best absolute pose estimation.} However, this configuration suffers from a low \highlight{SR and poor mRPE,} suggesting instability. In contrast, the T265 camera shows \highlight{better mRPE and a much higher SR, although it exhibits a higher mATE.}
For mono-inertial setups, depicted in Table~\ref{table:Mono-inertial_results}, the results are more varied. Some configurations fail to produce valid results, while others perform relatively well. OpenVINS with the T265 camera in the internal configuration stands out as one of the better-performing setups, showing solid results \highlight{in both mATE and mRPE}. 
In stereo configurations (see Table~\ref{table:stereo_results}), most setups \highlight{experience significant difficulties in achieving low mATE values, while the mRPE results remain largely comparable. ORB-SLAM3 stands out by attaining better consistency and higher overall accuracy in both mATE and mRPE.}
When adding inertial data to stereo configurations the results become more stable. \highlight{Both OpenVINS, which performs well with both internal and external T265 configurations, and SVO Pro using an external IMU exhibit comparable performance. ORB-SLAM3 with the external IMU stands out as the best performing method by achieving the lowest mATE and mRPE, but its relatively low SR makes it a rather unstable method.} Overall, stereo-inertial setups offer more reliable localization performance than pure stereo configurations.
The results for the RGBD configuration are presented in Table~\ref{table:RGBD_results}, demonstrating a significant improvement in performance. The inclusion of depth data enhances SLAM performance considerably, \highlight{with RGBD configurations showing good localization results in terms of both mATE and mRPE, and ORB-SLAM3 achieving a high SR, which makes it one of the most robust options tested.}
However, when inertial data is added to RGBD configurations, the results do not improve further; in fact, the performance tends to decline. Both the accuracy and \highlight{SR} drop compared to RGBD-only configurations, suggesting that the addition of inertial data, in this case, does not contribute positively to the results.

\begin{table*}[!ht]
\centering
\caption{\highlight{Performance of the best-performing sensor configuration for each SLAM method across different seasons, with OV representing OpenVINS, VF for VINS-Fusion, ORB for ORB-SLAM3, SVO for SVO Pro, DROID for DROID-SLAM, and DPV for DPV-SLAM. Sensor configurations are denoted as follows: S-I for stereo-inertial, D for RGBD, M for mono, I for internal IMU, and E for external IMU.}}
\resizebox{\textwidth}{!}{%
\begin{tabular}{l ccc ccc ccc ccc ccc}
    \toprule
    \textbf{SLAM}
    & \multicolumn{3}{c}{\textbf{Summer}}
    & \multicolumn{3}{c}{\textbf{Autumn}}
    & \multicolumn{3}{c}{\textbf{Winter}}
    & \multicolumn{3}{c}{\textbf{Spring}}
    & \multicolumn{3}{c}{\textbf{Overall}} \\
    \cmidrule(lr){2-4}
    \cmidrule(lr){5-7}
    \cmidrule(lr){8-10}
    \cmidrule(lr){11-13}
    \cmidrule(lr){14-16}
    & mATE & mRPE & SR
    & mATE & mRPE & SR
    & mATE & mRPE & SR
    & mATE & mRPE & SR
    & mATE & mRPE & SR \\
    \midrule
    
    OV (S-I T265 I)
    & \textbf{1.59} & \textbf{0.17} & 100.00
    & \textbf{1.00} & 0.33 & 80.00
    & 1.79 & 0.28 & 100.00
    & 1.34 & 0.17 & 100.00
    & 1.43 & 0.23 & 94.74 \\
    
    VF (S-I T265 I)
    & 4.21 & 0.82 & 80.00
    & 2.31 & 0.40 & 60.00
    & 2.83 & 0.56 & 100.00
    & 2.93 & 0.40 & 60.00
    & 3.13 & 0.57 & 73.68 \\
    
    ORB (D D435i)
    & 2.78 & 0.19 & 100.00
    & 1.06 & 0.17 & 100.00
    & 0.42 & 0.16 & 75.00
    & \textbf{0.46} & 0.17 & 100.00
    & \textbf{1.26} & \textbf{0.17} & 94.74 \\
    
    SVO (S-I T265 E)
    & 2.11 & 0.18 & 100.00
    & 1.50 & \textbf{0.16} & 100.00
    & 1.32 & 0.14 & 100.00
    & 1.81 & 0.17 & 100.00
    & 1.70 & 0.16 & 100.00 \\
    
    DROID (D D435i)
    & 3.87 & 0.43 & 80.00
    & 4.42 & 0.24 & 100.00
    & \textbf{0.34} & \textbf{0.07} & 100.00
    & 0.56 & \textbf{0.08} & 100.00
    & 2.32 & 0.21 & 94.74 \\
    
    DPVO (M D435i)
    & 6.40 & 2.13 & 100.00
    & 6.53 & 2.04 & 100.00
    & 6.48 & 2.02 & 100.00
    & 6.45 & 2.39 & 100.00
    & 6.47 & 2.15 & 100.00 \\
    
    DPV (M D435i)
    & 6.27 & 1.67 & 100.00
    & 6.39 & 1.93 & 100.00
    & 5.81 & 1.87 & 100.00
    & 6.41 & 1.88 & 100.00
    & 6.24 & 1.84 & 100.00 \\
    
    \cmidrule(lr){1-16}
    Mean
    & 3.89 & 0.80 & 94.29
    & 3.32 & 0.75 & 91.43
    & 2.71 & 0.73 & 96.43
    & 2.85 & 0.75 & 94.29
    & 3.22 & 0.76 & 93.99 \\

\bottomrule
\end{tabular}
}
\label{table:season_results}
\end{table*}

\begin{table*}[!ht]
\centering
\caption{\highlight{Performance of the best-performing sensor configuration for each SLAM method across different lighting conditions.}}
\resizebox{\textwidth}{!}{%
\begin{tabular}{l ccc ccc ccc ccc ccc}
    \toprule
    \textbf{SLAM} 
    & \multicolumn{3}{c}{\textbf{Day}} 
    & \multicolumn{3}{c}{\textbf{Dusk}}
    & \multicolumn{3}{c}{\textbf{Night}}
    & \multicolumn{3}{c}{\textbf{Night + Light}}
    & \multicolumn{3}{c}{\textbf{Overall}} \\
    \cmidrule(lr){2-4} 
    \cmidrule(lr){5-7} 
    \cmidrule(lr){8-10}
    \cmidrule(lr){11-13}
    \cmidrule(lr){14-16}
    & mATE & mRPE & SR
    & mATE & mRPE & SR
    & mATE & mRPE & SR
    & mATE & mRPE & SR
    & mATE & mRPE & SR \\
    \midrule
    
    OV (S-I T265 I)
    & 2.26 & 0.17 & 100.00 
    & \textbf{1.25} & \textbf{0.15} & 100.00
    & \textbf{1.53} & 0.19 & 100.00
    & 2.53 & 0.40 & 100.00
    & \textbf{1.89} & 0.23 & 100.00 \\
    
    VF (S-I T265 I)
    & 2.57 & 0.50 & 80.00
    & 3.83 & 0.54 & 60.00
    & 19.40 & 0.75 & 60.00
    & 3.95 & 0.98 & 60.00
    & 7.06 & 0.68 & 65.00 \\
    
    ORB (D D435i)
    & 0.47 & 0.17 & 100.00
    & 6.17 & 0.85 & 100.00
    & 8.68 & 0.17 & 60.00
    & 4.78 & 0.15 & 80.00
    & 4.61 & 0.36 & 85.00 \\
    
    SVO (S-I T265 E)
    & 1.54 & 0.16 & 100.00
    & 1.77 & 0.19 & 100.00
    & 2.65 & 0.21 & 80.00
    & 8.13 & \textbf{0.10} & 80.00
    & 3.31 & 0.17 & 90.00 \\
    
    DROID (D D435i)
    & \textbf{0.32} & \textbf{0.07} & 100.00
    & 1.28 & 0.18 & 100.00
    & 5.62 & \textbf{0.15} & 100.00
    & \textbf{0.51} & 0.11 & 100.00
    & 1.93 & \textbf{0.13} & 100.00 \\
    
    DPVO (M D435i)
    & 5.49 & 2.09 & 100.00
    & 5.90 & 2.88 & 100.00
    & 7.05 & 4.11 & 100.00
    & 5.35 & 1.98 & 100.00
    & 5.95 & 2.77 & 100.00 \\
    
    DPV (M D435i)
    & 5.47 & 1.80 & 100.00
    & 5.62 & 2.75 & 100.00
    & 7.15 & 3.33 & 100.00
    & 4.69 & 1.51 & 100.00
    & 5.73 & 2.35 & 100.00 \\
    
    \cmidrule(lr){1-16}
    Mean
    & 2.59 & 0.71 & 97.14
    & 3.69 & 1.08 & 94.29
    & 7.44 & 1.27 & 85.71
    & 4.28 & 0.76 & 88.57
    & 4.35 & 0.96 & 91.43 \\

\bottomrule
\end{tabular}
}
\label{table:lighting_results}
\end{table*}

\begin{figure*}[h!]
  \centering
  \includegraphics[width=0.985\linewidth]{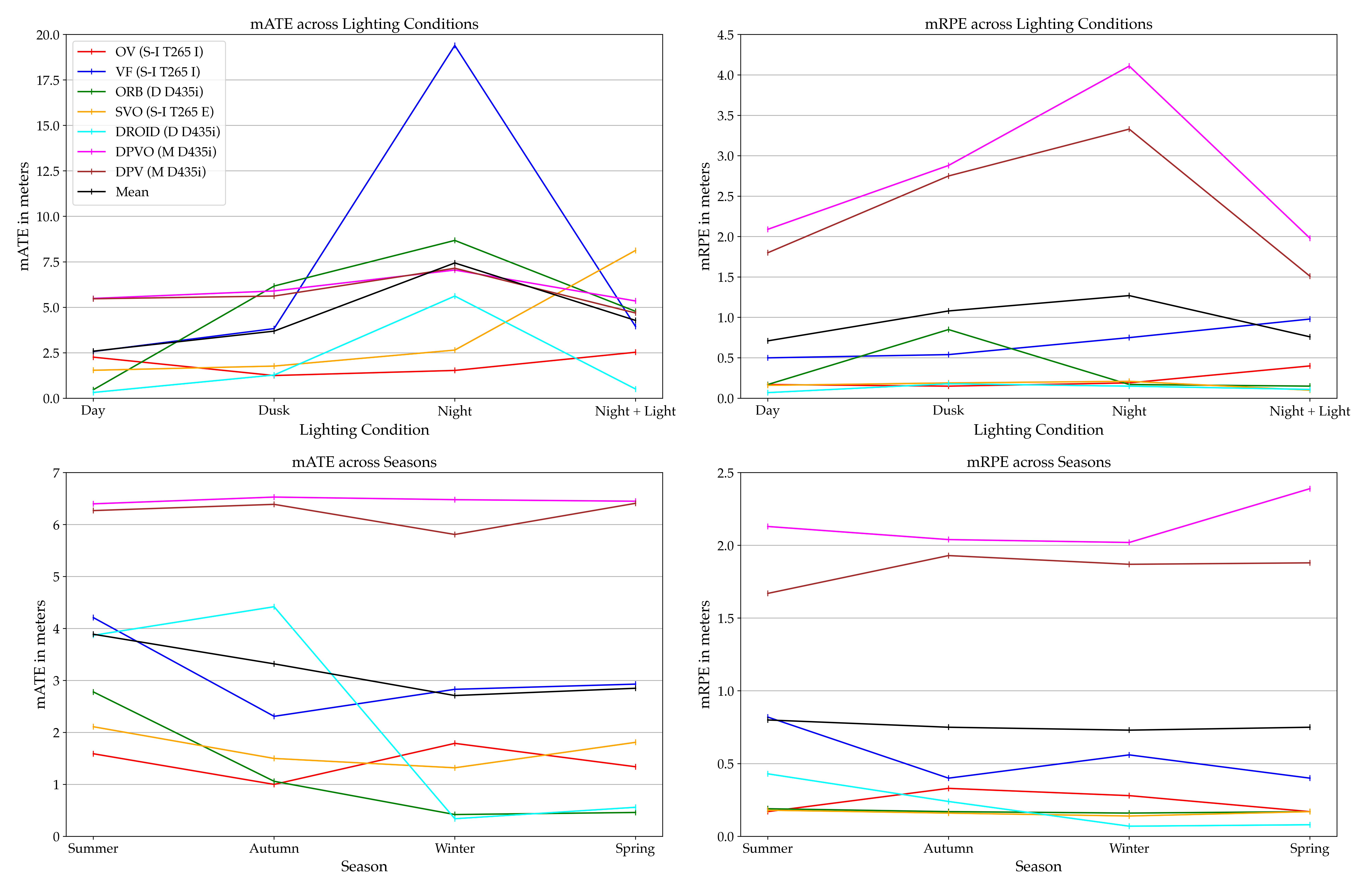}
  \caption{\highlight{Comparison of best-performing SLAM methods regarding mATE and mRPE under different lighting conditions and seasons.}\label{fig:benchmark_overview}}
\end{figure*}

\begin{figure*}[b!]
  \centering
  \footnotesize
  \setlength{\tabcolsep}{0.05cm}
  
  {\renewcommand{\arraystretch}{1}
    \begin{tabular}{p{6.5cm} p{5.0cm} p{6.5cm}}
        \includegraphics[height=6cm]{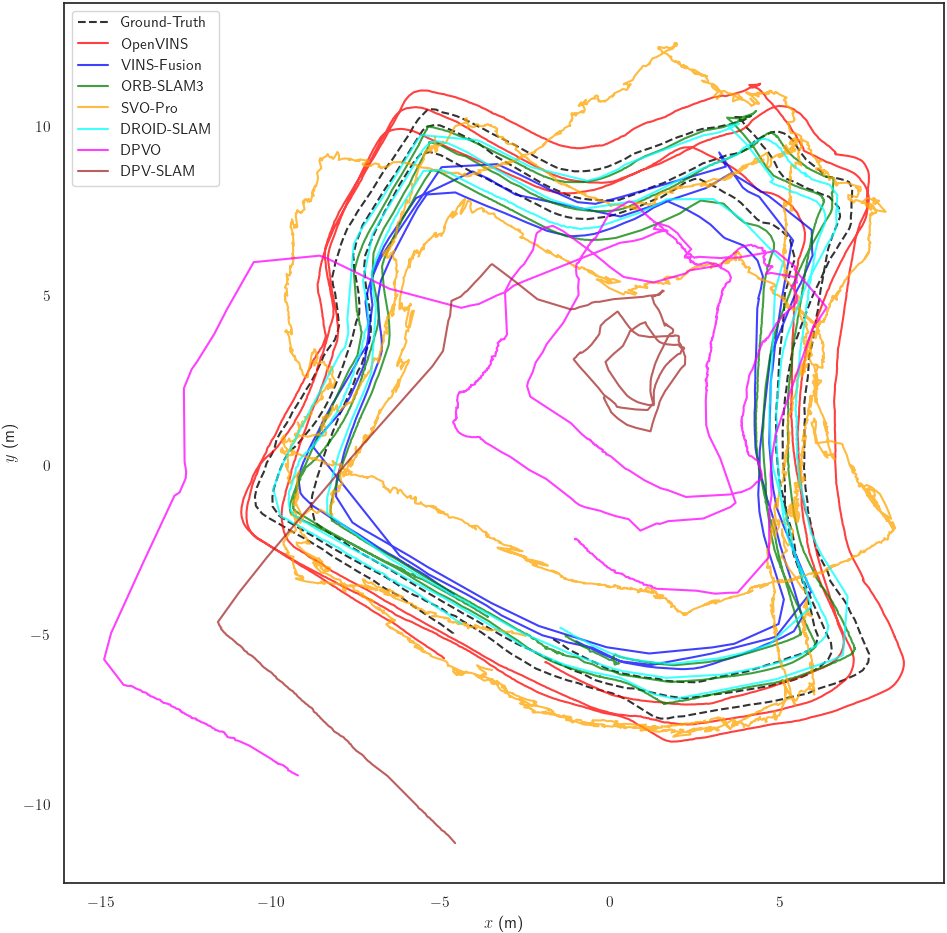} &
        \includegraphics[height=6cm]{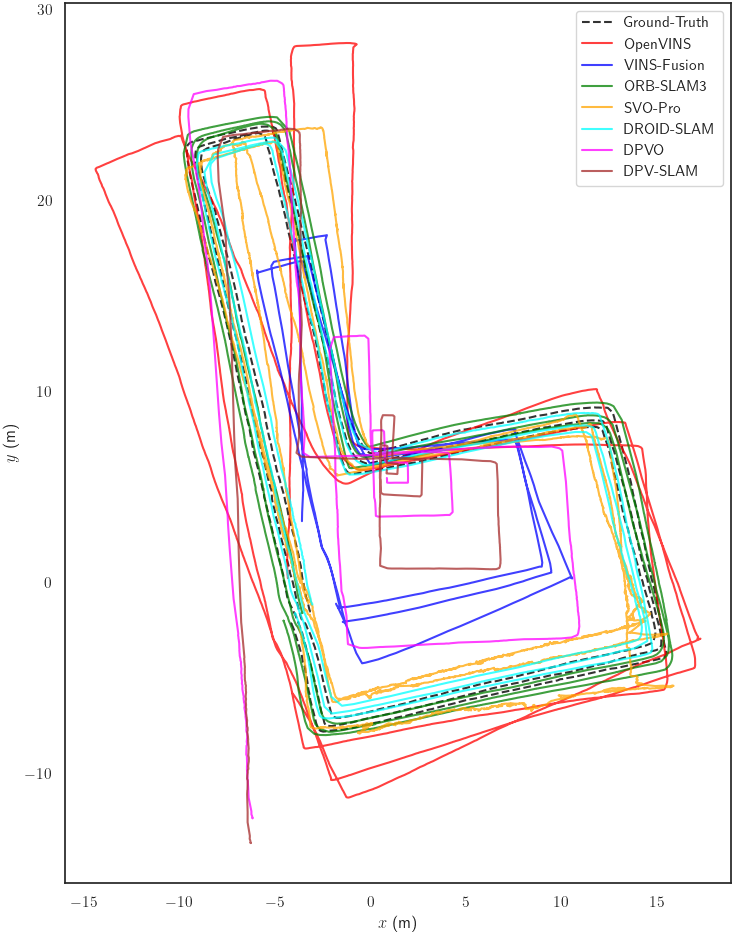} &
        \includegraphics[height=6cm]{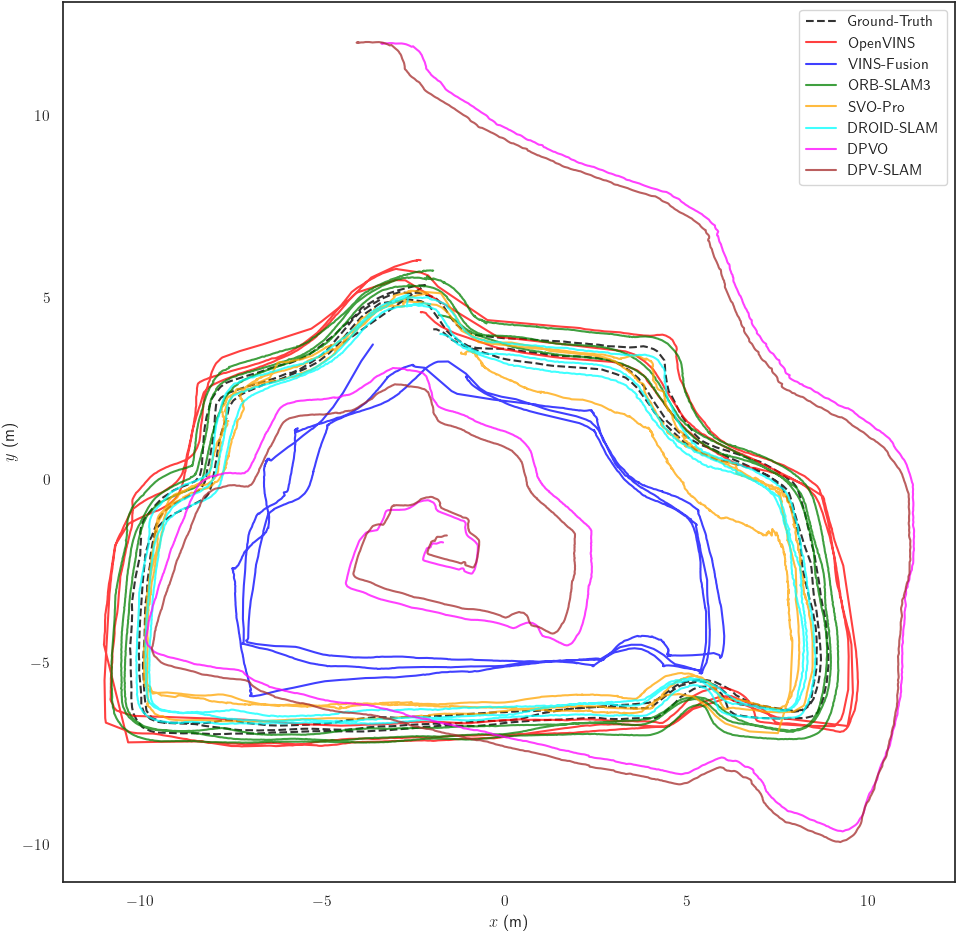} \\
        \multicolumn{1}{c}{(a) Park} & \multicolumn{1}{c}{(b) Campus Large} & \multicolumn{1}{c}{(c) Garden Large} \\[0.5em]
    \end{tabular}}

    \vspace{0.25em}
  
  {\renewcommand{\arraystretch}{1}
    \begin{tabular*}{0.98\textwidth}{@{}l@{\extracolsep{\fill}}r@{}}
        \includegraphics[height=6cm]{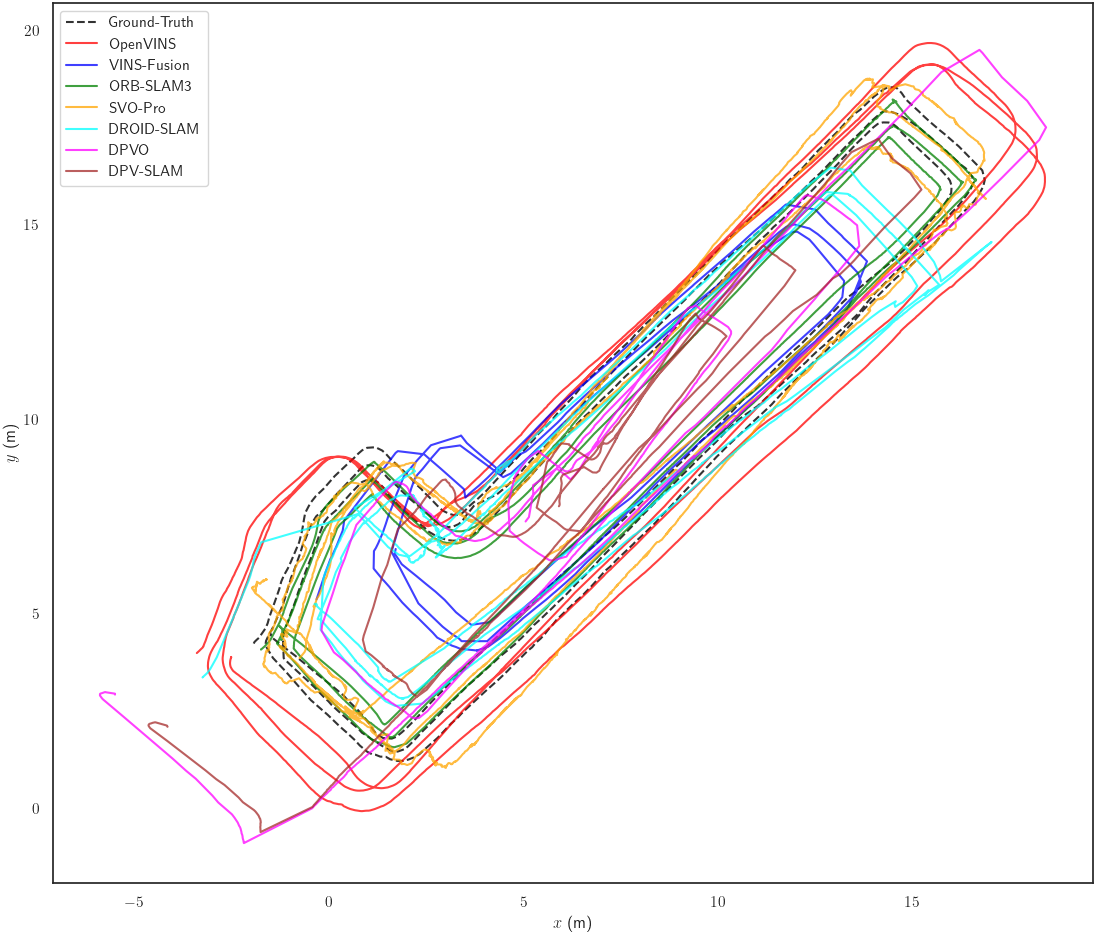} &
        \includegraphics[height=6cm]{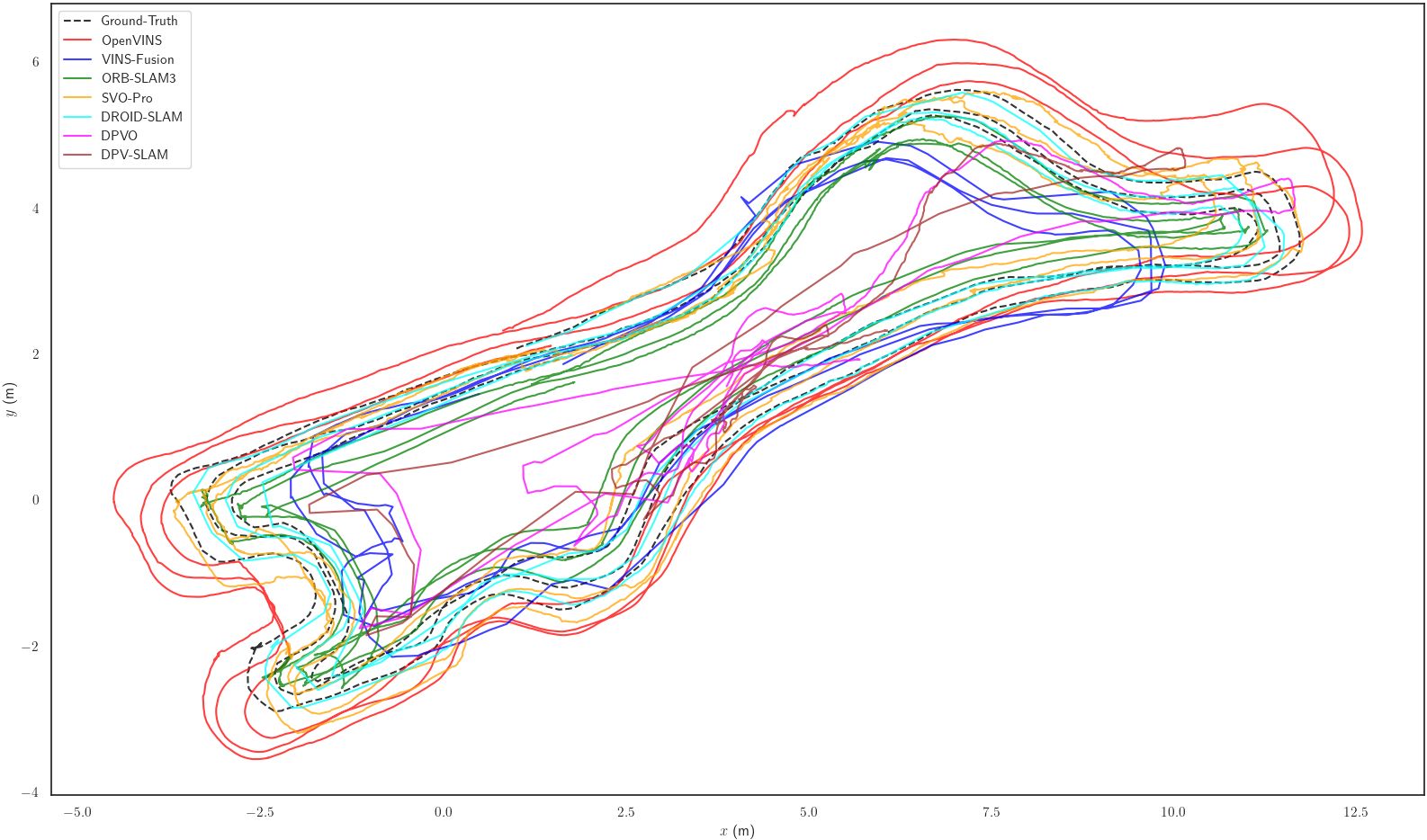} \\
        \multicolumn{1}{c}{(d) Campus Small} & \multicolumn{1}{c}{(e) Garden Small} \\[0.5em]
    \end{tabular*}}
    
    \caption{Qualitative trajectory comparisons across environments. (a) Park, (b) Campus Large, (c) Garden Large, (d) Campus Small, and (e) Garden Small. These figures illustrate the performance and stability of SLAM algorithms under various spatial complexities and environmental conditions, showing variations in drift, trajectory consistency, and overall accuracy.}
    \label{fig:qual_res_trajectories}
\end{figure*}

In addition to the overall performance evaluation of the SLAM algorithms, we also examine the impact of environmental factors, including seasons and lighting conditions, on the best-performing configurations. 
Seasonal changes play a critical role in the performance of SLAM systems, as shown in Table~\ref{table:season_results} and visualized in Figure~\ref{fig:benchmark_overview}. Winter and spring generally offer the most favorable conditions for accurate localization, with lower vegetation density and more stable lighting reducing visual complexity. In contrast, summer and autumn, characterized by denser vegetation and more variable lighting, lead to significantly higher \highlight{mATE} values. \highlight{While the absolute errors (mATE) are clearly affected by seasonal variations, the relative errors (mRPE) remains only minimally impacted.} Systems such as ORB-SLAM3 and DROID-SLAM perform well during winter and spring, but encounter increased \highlight{mATEs} in summer, reflecting the seasons' characteristics. Notably, OpenVINS (stereo-inertial T265 - internal) demonstrates strong year-round consistency, while ORB-SLAM3 and DROID-SLAM (RGBD D435i) excel in winter and spring.
Lighting conditions also have a profound impact on SLAM performance. 

The results show that SLAM systems perform most reliably in daylight, where they achieve the lowest \highlight{mATE} and the highest \highlight{SR}, as shown in Table~\ref{table:lighting_results}. For example, DROID-SLAM (RGBD D435i) consistently performs well during daytime across various locations, while OpenVINS (stereo-inertial T265 internal) also exhibits stable performance in both daylight and dusk conditions. As light decreases in dusk and night scenarios, \highlight{both mATE and mRPE} performance declines sharply for most algorithms, with night conditions posing the greatest challenge -- though OpenVINS shows relative resilience. Many SLAM methods face difficulties maintaining accuracy in low-light environments, as evidenced by higher error rates and lower \highlight{SR} at night. ORB-SLAM3 (RGBD D435i), for instance, shows a sharp increase in ATE during night conditions despite performing well during the day. Even when external lighting is introduced, as in the night + light scenario, the improvement is only marginal, highlighting the need for enhanced robustness under low-visibility conditions.  These results indicate that while some configurations, like DROID-SLAM and OpenVINS, are more resilient under suboptimal lighting, many SLAM methods still struggle with low-light feature extraction, especially in larger environments.

The qualitative results, depicted in Figure~\ref{fig:qual_res_trajectories}, reveal key performance variations across different SLAM configurations and environments. In general, ORB-SLAM3 and DROID-SLAM exhibit strong results in locations such as the Park, Campus Small, and Garden Large, handling visual cues effectively and maintaining stability. OpenVINS and VINS-Fusion, however, encounter scaling issues in multiple environments, particularly the Park and Campus Large, leading to less consistent performance. Similarly, SVO Pro faces challenges with drift and scale issues, resulting in jittery trajectories, especially in larger, more complex settings. Monocular approaches like DPVO and DPV-SLAM struggle significantly with drift and scaling issues across most environments, especially in the open or vegetative areas of Campus Large and Garden Small, due to inherent scale ambiguity. 

\subsection{Discussion}

The results demonstrate that current SLAM systems face significant challenges in generalizing across the diverse environmental conditions in this dataset, such as seasonal and lighting variations, revealing substantial gaps in robustness and adaptability. We hope to facilitate further research on those shortcomings with this study.

Seasonal changes introduce substantial variability in SLAM performance, particularly in summer and autumn when denser vegetation and shifting lighting make feature extraction difficult. Algorithms such as ORB-SLAM3 and DROID-SLAM perform better during winter and spring but experience notable declines in accuracy during summer (refer to Table~\ref{table:season_results}). These seasonal patterns suggest that current SLAM systems are not yet optimized for environments where visual complexity fluctuates due to natural factors such as vegetation growth.

Lighting conditions further exacerbate these challenges. While SLAM systems perform well in daylight, as evidenced by the low \highlight{mATEs, low mRPEs and high SR}, performance sharply declines in low-light scenarios such as dusk and night. Even with external lighting, many methods still struggle to maintain accuracy, as shown by the increase in \highlight{mATE and mRPE} values at night (refer to  Table~\ref{table:lighting_results}). This indicates that the feature extraction process in low-light environments remains a significant bottleneck for SLAM systems. Qualitatively, this is evident in locations like Campus Large and Park, where even with controlled environments, systems like SVO Pro exhibit drift and jitter under suboptimal lighting, contributing to higher error rates.

The camera and sensor configurations also play a crucial role in determining SLAM accuracy. Mono cameras, for instance, tend to perform adequately in smaller environments such as Garden Small but falter in larger, more complex settings. Monocular approaches, such as DPVO and DPV-SLAM, heavily struggle with scaling issues due to the inherent scale ambiguity of these systems, particularly in larger locations such as Campus Large and Garden Large, where visual cues are more difficult to anchor (refer to  Figure~\ref{fig:qual_res_trajectories}). In contrast, RGBD configurations such as ORB-SLAM3 (RGBD D435i) benefit from depth data, leading to improved performance in both small and large environments. However, adding inertial data does not consistently improve performance. In some cases, as seen with RGBD-inertial setups, inertial data even reduces accuracy and success rates, suggesting that the integration of visual and inertial data requires further refinement.

The results of this evaluation underscore the need for more robust SLAM systems capable of adapting to the unpredictable dynamics of outdoor environments. While some configurations, such as OpenVINS (stereo-inertial T265 internal) and DROID-SLAM (RGBD D435i), show promising results, many algorithms still struggle with low-light conditions, high vegetation density, and large, unstructured environments. This dataset reveals critical challenges that SLAM systems must overcome to achieve long-term, robust localization and mapping in natural outdoor settings. Further research is needed to enhance the resilience of the SLAM algorithms to environmental variability and improve their adaptability across diverse scenarios.

\section{Conclusion}
\label{sec:conclusions}

In this paper, we presented a large-scale benchmark dataset designed to evaluate SLAM algorithms under a variety of spatial and environmental conditions, allowing the assessment of complex real-world scenarios. The benchmark results demonstrate that while some SLAM configurations, particularly those using RGBD or stereo-inertial setups, show promising performance in structured environments, challenges remain significant under varying seasonal and lighting conditions. Notably, performance degrades in low-light and high-vegetation settings, with substantial increases in trajectory errors during summer and autumn, where denser vegetation and variable lighting complicate feature extraction.

Lighting conditions further impact robustness, as most SLAM methods achieve reliable results in daylight but experience sharp declines in accuracy at dusk and night. Despite improvements with external lighting, many configurations, especially monocular approaches, continue to struggle with maintaining accuracy due to scale ambiguity and drift. The qualitative analysis underscores that algorithms such as ORB-SLAM3 and DROID-SLAM show resilience in structured or well-lit conditions but face limitations in environments with limited visual features or uneven lighting, such as rooftop spaces or shaded garden areas.
These results highlight that while current SLAM methods have made strides, their robustness in highly variable \highlight{semi-structured} natural environments is limited, especially regarding long-term mapping and consistency. Further development is needed to enhance SLAM resilience in outdoor settings, and this dataset provides a comprehensive foundation for addressing these gaps.

\section*{Acknowledgments}
We thank Joshua Uhl for his support during the data capturing phase and Sabrina Kaniewski for her valuable assistance with reviewing and editing.

\footnotesize
\bibliographystyle{IEEEtran}
\bibliography{bibliography}

\begin{IEEEbiography}[{\includegraphics[width=1in,height=1.25in,clip,keepaspectratio]{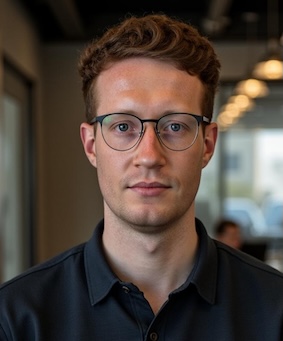}}]{Fabian Schmidt}
received his M.S. degree in Applied Computer Science from Esslingen University of Applied Sciences in 2023 and is currently a Ph.D. student supervised by Abhinav Valada and Markus Enzweiler. His research focuses on visual SLAM and autonomous navigation for mobile robots in unstructured outdoor environments.
\end{IEEEbiography}

\begin{IEEEbiography}[{\includegraphics[width=1in,height=1.25in,clip,keepaspectratio]{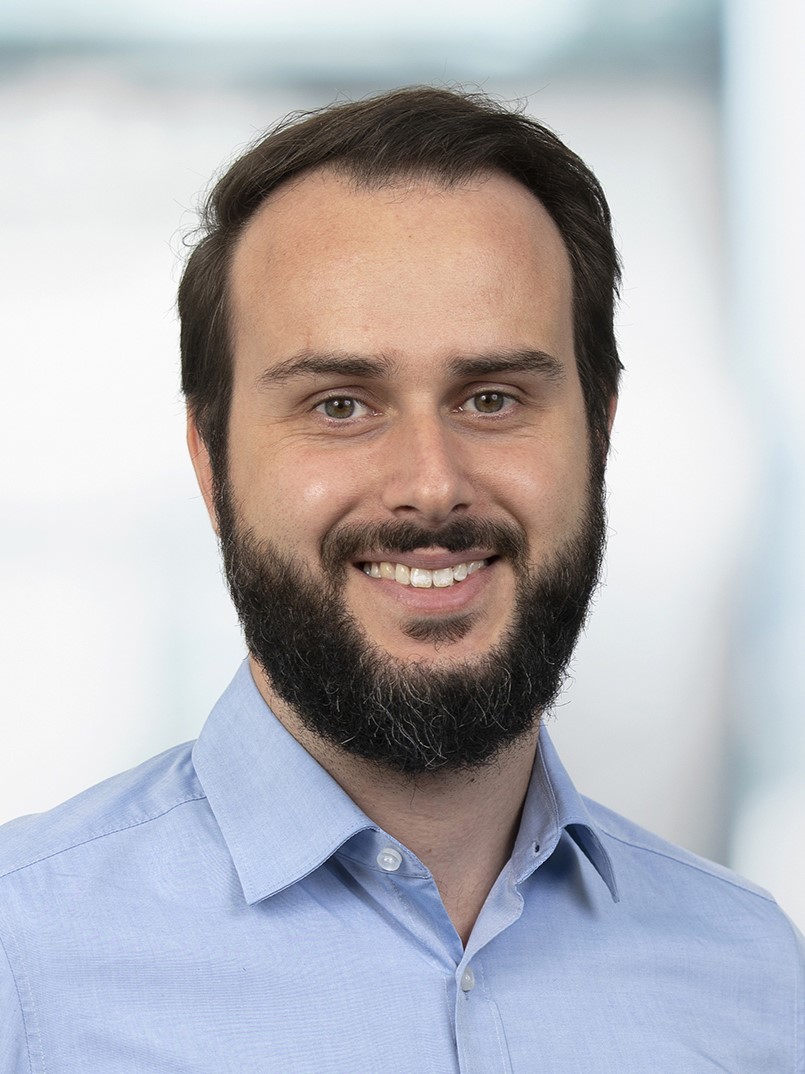}}]{Julian Daubermann}
received the M.S. degree in Computer Science from Hochschule Mannheim in 2012. From 2013 to 2018, he was a doctoral researcher at the European Innovation and Technology Center of John Deere in Kaiserslautern, where he worked on autonomous agricultural machines. He received the Ph.D. degree in Engineering from Technical University Berlin in 2023. Since 2018, he has been working at STIHL in pre-development for robotics and autonomous systems, focusing on environmental sensing.
\end{IEEEbiography}

\begin{IEEEbiography}[{\includegraphics[width=1in,height=1.25in,clip,keepaspectratio]{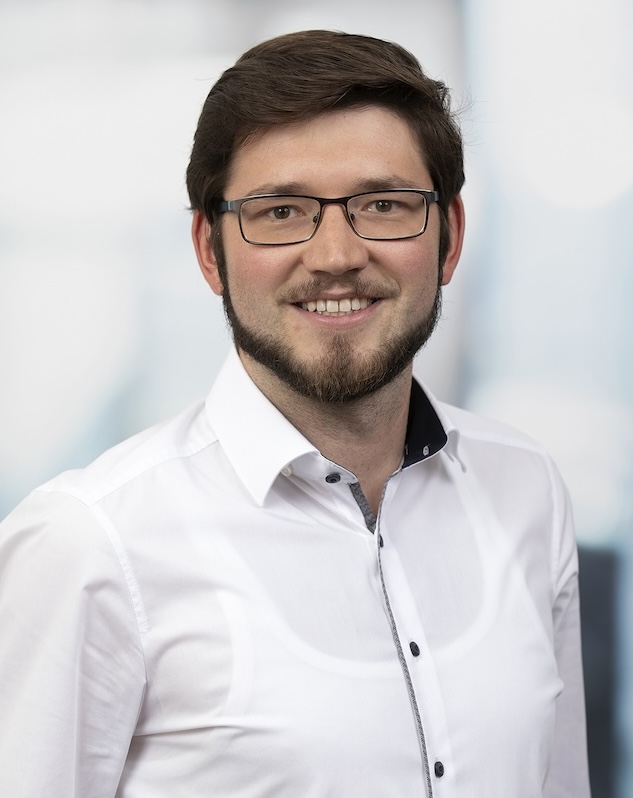}}]{Marcel Mitschke}
received his M.S. degree in Mechatronics from the University of Stuttgart and M.S. degree in Mechanical Engineering from the Toyohashi University of Technology in the year 2017. His research focused on path planning and environmental sensing for mobile robots. From 2018 he is working in the robotics research and development department of STIHL.
\end{IEEEbiography}

\begin{IEEEbiography}[{\includegraphics[width=1in,height=1.25in,clip,keepaspectratio]{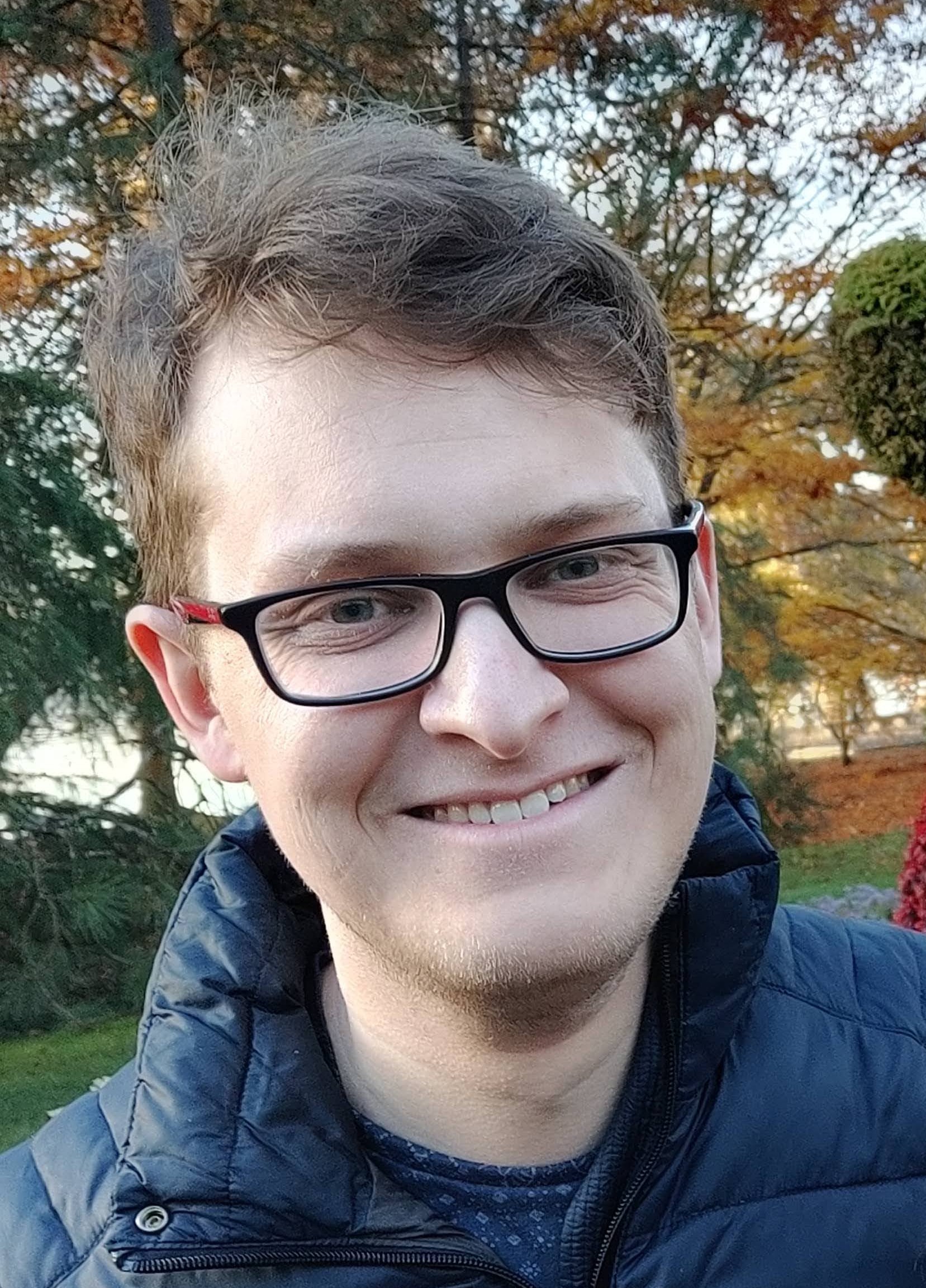}}]{Constantin Blessing}
received his M.S. degree in Applied Computer Science from Esslingen University of Applied Sciences in 2023 and is currently a Ph.D. student supervised by Markus Enzweiler. His research focuses on collaborative task accomplishment by multi-agent systems in industrial environments.
\end{IEEEbiography}

\begin{IEEEbiography}[{\includegraphics[width=1in,height=1.25in,clip,keepaspectratio]{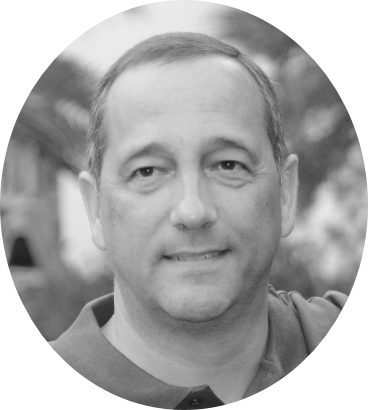}}]{Stephan Meyer.}
Stephan Meyer received his M.S. degree in Techno-Mathematics in 1996 and his Ph.D. in Mechanical Engineering in 2003, both from Technical University of Karlsruhe. In 2003, he started his industrial career as a CFD Engineer at ANDREAS STIHL AG in Waiblingen where he is still employed today. In 2006, he joined the pre-development department for gasoline powered chain saws with focus on motor management systems. In 2017, he built up and heads the pre-development group for STIHL’s autonomous system with localization and environmental perception being the main topics.
\end{IEEEbiography}

\begin{IEEEbiography}[{\includegraphics[width=1in,height=1.25in,clip,keepaspectratio]{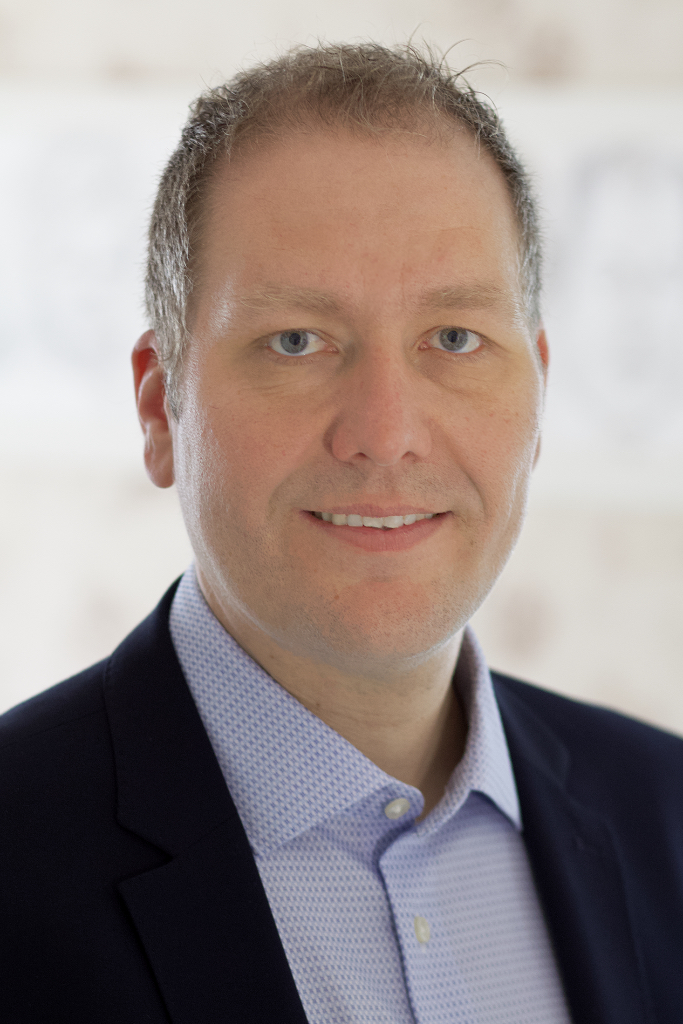}}]{Markus Enzweiler}
received the M.S. degree in Computer Science from the University of Ulm, in 2005. In 2002 and 2003 he was a visiting student researcher at the Centre for Vision Research at York University, Toronto, Canada. He received the Ph.D. degree in Computer Science from the University of Heidelberg in 2011. From 2010 to 2020, he has been working with Mercedes-Benz Research \& Development in Stuttgart focusing on camera- and LiDAR-based scene understanding for self-driving cars. Since 2021, he is a Full Professor of Computer Science at Esslingen University of Applied Sciences, where he founded and heads the Institute for Intelligent Systems. 
\end{IEEEbiography}

\begin{IEEEbiography}[{\includegraphics[width=1in,height=1.25in,clip,keepaspectratio]{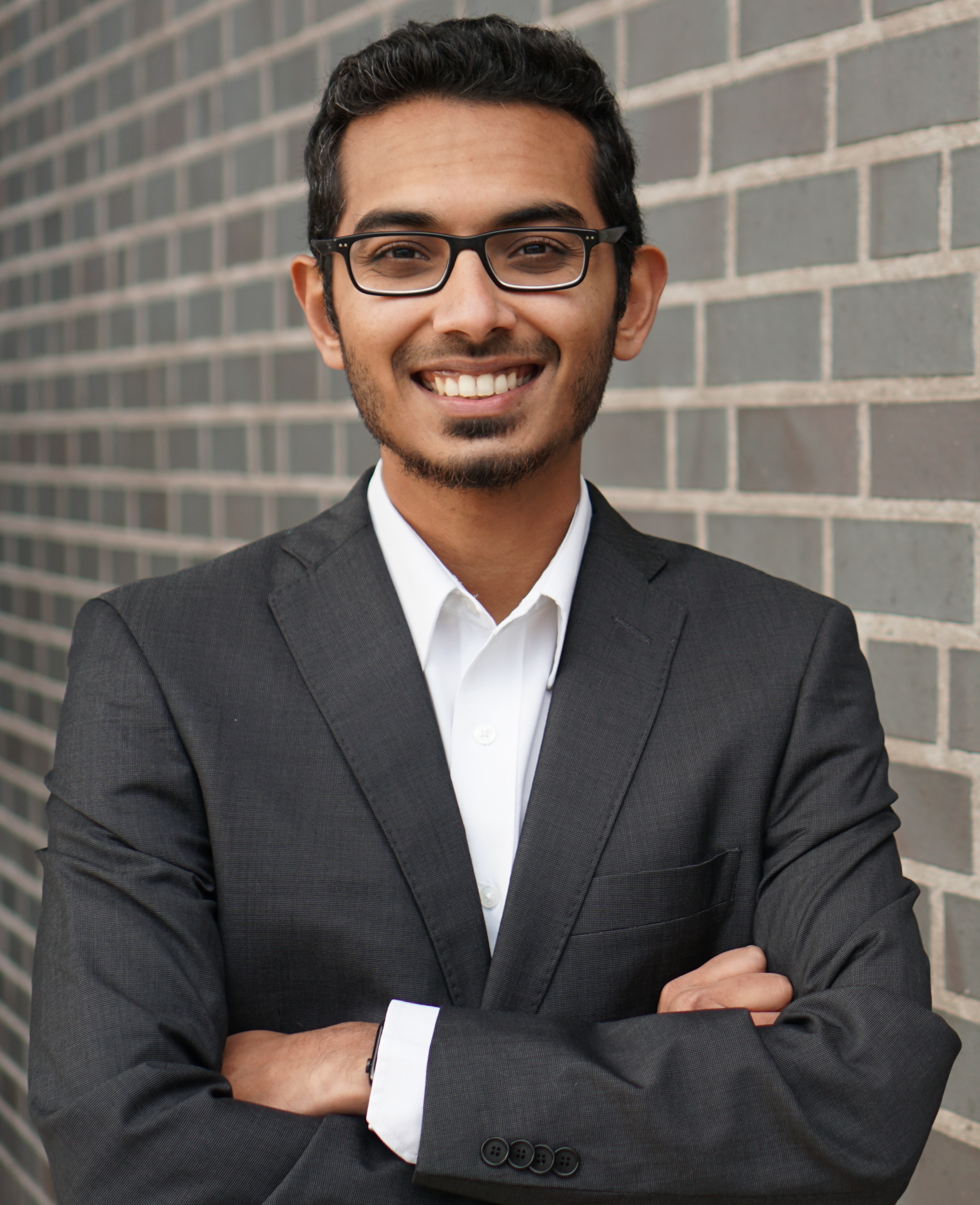}}]{Abhinav Valada}
is a Full Professor and Director of the Robot Learning Lab at the University of Freiburg, Germany. He is a member of the Department of Computer Science, a principal investigator at the BrainLinks-BrainTools Center, and a founding faculty of the European Laboratory for Learning and Intelligent Systems (ELLIS) unit at Freiburg. He received his Ph.D. in Computer Science from the University of Freiburg in 2019 and his M.S. degree in Robotics from Carnegie Mellon University in 2013. His research lies at the intersection of robotics, machine learning, and computer vision with a focus on tackling fundamental robot perception, state estimation, and planning problems using learning approaches in order to enable robots to reliably operate in complex and diverse domains. Abhinav Valada is a Scholar of the ELLIS Society, a DFG Emmy Noether Fellow, and co-chair of the IEEE RAS TC on Robot Learning.
\end{IEEEbiography}

\vspace{11pt}

\vfill

\end{document}